\pdfoutput=1

 \documentclass[11pt]{article}

\usepackage[]{EACL2023}

\usepackage{times}
\usepackage{latexsym}

\usepackage[utf8]{inputenc} 
\usepackage[T1]{fontenc}    

\usepackage{wrapfig,lipsum,booktabs}

\usepackage{url}
\usepackage{times}
\usepackage{latexsym}
\usepackage{todonotes}
\usepackage{mathtools}
\usepackage{tikz}
\usetikzlibrary{arrows,decorations.pathmorphing,fit,positioning}
\newcommand*\circled[1]{\tikz[baseline=(char.base)]{
            \node[shape=circle,draw,inner sep=1pt] (char) {#1};}}

\newcommand{\blue}[1]{\textcolor{blue}{#1}\color{blue}\xspace}
\newcommand{\red}[1]{\textcolor{red}{#1}\color{red}\xspace}
\usepackage{amssymb}
\usepackage{lipsum}
\newcommand{\fsc}[1]{\textsc{#1}}
\usepackage{tabularx}

\usepackage{arydshln}

\newcommand{\st} {{\color{black}{t}}}
\newcommand{\bx}{\mathbf{x}}
\newcommand{\by}{\mathbf{y}}
\newcommand{\bz}{\mathbf{z}}
\newcommand{\bt}{{\color{black}\mathbf{t}}}
\newcommand{\mut}{\mathcal{I}}
\usepackage{afterpage}

\usepackage{times}
\usepackage{xspace}
\usepackage{latexsym}
\usepackage{booktabs}
\usepackage{amssymb}
\usepackage{pifont}
\usepackage{times}
\usepackage{latexsym}
\usepackage{times}
\usepackage{latexsym}
\usepackage{microtype}
\usepackage{multirow, multicol}
\usepackage{cite}
\usepackage{url}
\usepackage{makecell}

\usepackage{tabularx}
\usepackage{booktabs}
\usepackage{multirow}
\usepackage{microtype}
\usepackage{times}
\usepackage{latexsym}
\usepackage{amsmath}
\usepackage{amssymb}
\usepackage{graphicx}  
\usepackage{lipsum}
\usepackage{multirow}
\usepackage{tabularx}
\usepackage{arydshln}
\usepackage{afterpage}
\usepackage{subcaption}
\usepackage{float}
\usepackage{bbm}
\usepackage{booktabs}

\usepackage{graphicx}  
\usepackage{subcaption}
\usepackage{amsfonts}
\usepackage{multicol}
\usepackage{amsmath}
\usepackage{multirow}
\usepackage{tikz}

\DeclarePairedDelimiterX{\infdivx}[2]{\big(}{\big)}{%
  #1\;\delimsize\|\;#2%
}
\newcommand{\infdiv}{\mathrm{KL}\infdivx}
\usepackage[inline,shortlabels]{enumitem}

\usepackage[nameinlink]{cleveref}
\creflabelformat{equation}{#1#2#3}
\crefname{equation}{Eq.}{Eqs.}  
\Crefname{equation}{Eq.}{Eqs.}
\crefname{figure}{Fig.}{Figs.}  
\Crefname{figure}{Fig.}{Figs.}
\crefname{appendix}{Appendix}{App.}
\Crefname{appendix}{Appendix}{App.}

  \newcommand{\argmax}{\operatornamewithlimits{argmax}}

\usepackage{amsthm}
\usepackage{microtype}

\newtheorem*{definition}{Definition}

\newtheorem{thm}{Theorem}[]

\newtheorem*{thmwonumb}{Theorem}
\crefname{thm}{Theorem}{}
\crefname{prop}{Proposition}{}
\crefname{cor}{Corollary}{}

\usepackage{url}            
\usepackage{booktabs}       
\usepackage{amsfonts}       
\usepackage{nicefrac}       
\usepackage{microtype}      
\usepackage{xcolor}         
\usepackage{wrapfig}
\newcommand\norm[1]{\left\lVert#1\right\rVert}
\usepackage{bbm}

\usepackage{algorithm,algorithmicx,algpseudocode}
\newcommand*\myat{{\fontfamily{ptm}\selectfont\small @}}

\algrenewcommand\algorithmicfunction{\textbf{Function}}

\title{RevUp: Revise and Update Information Bottleneck\\for Event Representation}
\author{Mehdi Rezaee ~~~ Francis Ferraro\\
  Department of Computer Science\\
  University of Maryland Baltimore County\\
  Baltimore, MD 21250 USA \\
{\tt \{\href{mailto:rezaee1@umbc.edu}{rezaee1}, \href{mailto:ferraro@umbc.edu}{ferraro}\}\myat umbc.edu}
}

\begin{document}

\maketitle

\begin{abstract}
The existence of external (``side'') semantic knowledge has been shown to result in more expressive computational event models. %
To enable the use of side information that may be noisy or missing, we propose a semi-supervised information bottleneck-based discrete latent variable model. %
We reparameterize the model's discrete variables with auxiliary continuous latent variables and a light-weight hierarchical structure. %
Our model is learned to minimize the mutual information between the observed data and optional side knowledge that is not already captured by the new, auxiliary variables. %
We theoretically show that our approach generalizes past approaches, and perform an empirical case study of our approach on event modeling. 
We corroborate our theoretical results with strong empirical experiments, showing that the proposed method outperforms previous proposed approaches on multiple datasets.

\end{abstract}

\section{Introduction} 
\label{sec:introduction}
In this work, we are interested in addressing limitations in how computational event modeling can make use of relevant, supplementary semantic knowledge. %
This is because when modeling text descriptions of complex situations, such as newspaper descriptions of real world events, learning how to encode richer information about those descriptions can be a fruitful way of improving modeling performance~\citep{judea-strube-2015-event,xia-etal-2021-lome}.
E.g., if we are dealing with sequences of events, like a newspaper report of a stock or commerce transaction, then being able to encode that ``buying'' or ``selling,'' even though they may have nuanced connotation differences, are instances of the same general event (a \textsc{transaction} event) can improve downstream predictive performance on what events might happen next~\citep{ferraro-2016-upf,rezaee2021event}. %

\begin{figure}[t]
\centering
\includegraphics[width=0.35\textwidth]{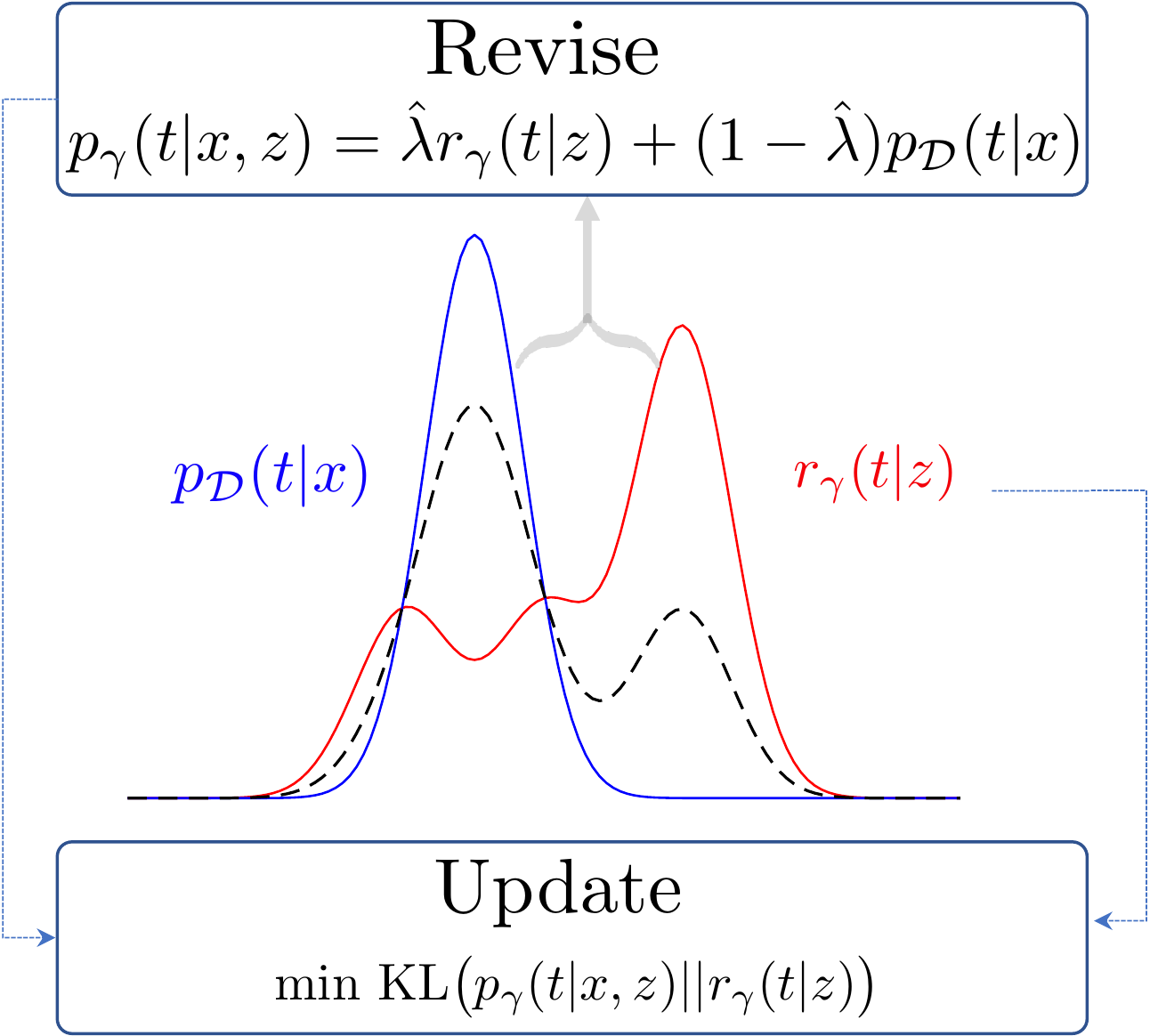}
\caption{Revise and Update steps for an observed node. We update the proposed distribution {\red{$r_{\gamma}(t\vert z)$}}  with the empirical distribution {\blue{$p_{\mathcal{D}}(t\vert x)$}} to produce the revised distribution $p_{\gamma}(t\vert x,z)$ (dashed). 
We then minimize the KL divergence between the proposed and revised distributions to update the proposed distribution. For unsupervised nodes, we rely on {\red{$r_{\gamma}(t\vert z)$}} without updating.
}
\label{fig:RevUpSteps}
\end{figure}

However, there is not an obvious single way to learn to encode this richer information, as three different questions naturally come to mind:
\begin{enumerate*}[(1)]
\item If the model is representationally limited, can we address these representational limitations of the model itself, such as through developing richer latent representations $\bz$ of the input $\bx$?
\item If there is available background or side information $\bt$ that may be especially relevant for the modeling task at hand, can we develop systems that make use of it? %
\item Even when side information is available, it may be noisy, e.g., it may not always be available (missing data) or it may contain errors: how can we make our models robust to this noisy side information? %
\end{enumerate*}  
In the context of a text-based sequence modeling problem, we propose an approach that addresses all three of these questions. %

\begin{figure*}
   \centering
   \includegraphics[width=0.95\linewidth]{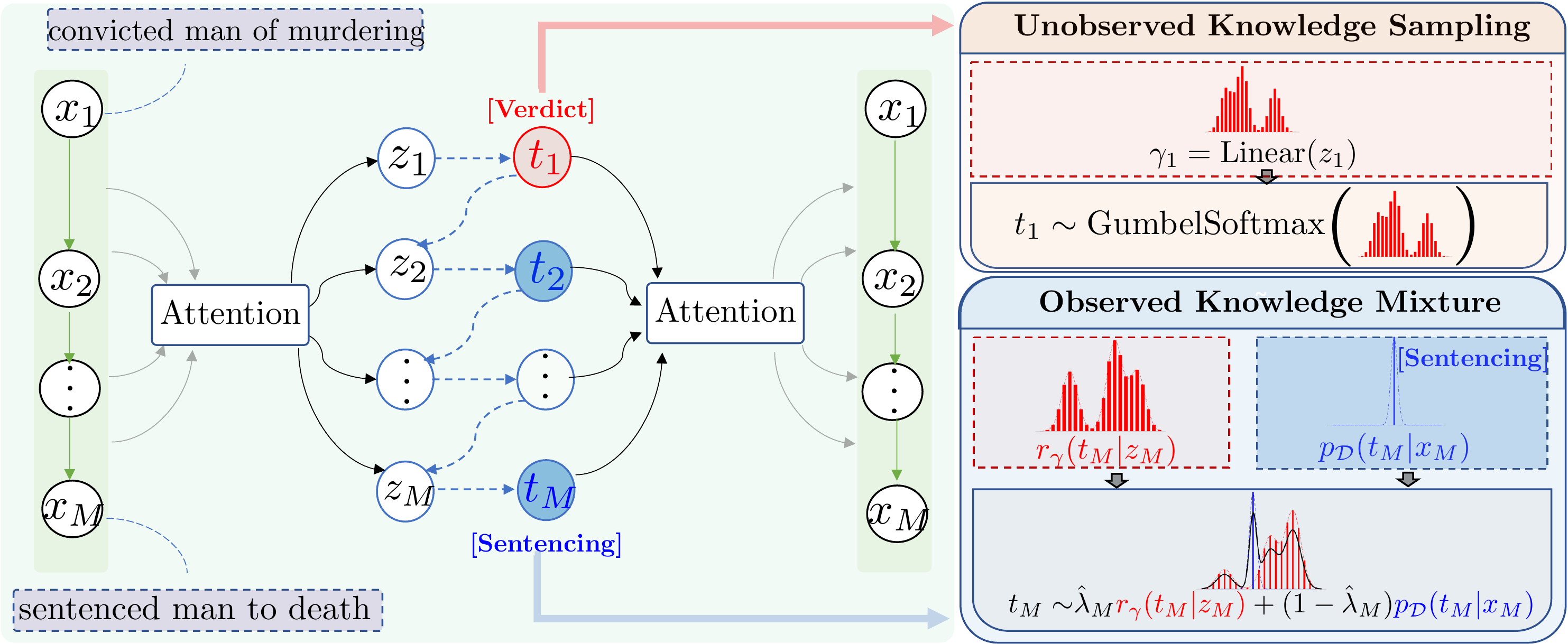}
   \caption{
   Components of RevUp for event modeling. We encode the sequence of events $\bx$ into Gaussian latent $\bz$ and discrete knowledge $\bt$. Red nodes such as {\red{$t_1$}} are latent:
   {\red{\textsc{[Verdict]}}} for the \textit{convicted man of murdering} event. RevUp predicts these latent nodes by sampling from the GumbelSoftmax distribution.
   Blue nodes such as {\blue{$t_M$}} are observed:     {\blue{\textsc{[Sentencing]}}} for the \textit{sentenced man to death} event. We use this observed node to modify the proposed distribution and then we draw a sample from the revised distribution. 
   }
   \label{fig:eventArch}
\end{figure*}

We provide a conceptual overview of our method---RevUp---in \cref{fig:RevUpSteps}. %
The building blocks of RevUp are \textbf{revision} of modeling side information $\bt$: forming a new distribution by combining the (red) learned proposal distribution with (blue) empirical information about when particular aspects of side knowledge appeared in training, and \textbf{updating} by minimizing the KL-divergence between the new distribution with the previous proposed one. %

Our approach is inspired by recent work~\citep{kong2019mutual}, which argued that a key to the success of neural advances in NLP~\citep[i.a.,]{mikolov2013distributed,devlin2018bert,yang2019xlnet} is that they fall within the information theoretic-based InfoMax framework~\citep{linsker88Info}. In this viewpoint, neural advances can be attributed to implicitly maximizing the mutual information between different representations of the same document. %

However, we additionally want to use ``side'' or ``external'' knowledge to aid our modeling. 
The use of side information has long been desired for effective representation~\citep{wyner1975side,wyner1976rate}. %
While this side information can occur in a variety of forms~\citep{chen-etal-2020-trying,padia-2018-surface}, we broadly view it as a \textit{concise, abstract view of information} conveyed in the main input.

Our work centers on the idea that the latent neural representations $\bz$ and side knowledge $\bt$ act as complementary representations of the same input $\bx$: $\bz$ provides a compact representation of the input itself, while $\bt$ provides more generic information about the data or task. 
As such, we use an InfoMax-inspired formulation to incorporate external knowledge into the modeling and latent representation aspects. %
Since recent work \citep{joy2022learning,chen2019variational} has suggested that providing some sort of ``guidance'' to these latent variables is beneficial and can lead to these guided models outperforming both fully supervised and fully unsupervised methods, learning with less-than-full observation of structured side knowledge is a requirement. %
We illustrate this in \cref{fig:eventArch}, where the correct value for the first piece of side information ($t_1 = \mathrm{Verdict}$) is missing but the last piece of side information is known ($t_M = \mathrm{Sentencing}$). Our model must be able to operate over this partially observed sequence of side knowledge as a way of modeling the original event description. %

We propose a unified approach, RevUp (\textbf{Rev}ise and \textbf{Up}date Information Bottleneck), that maximizes the mutual information between the external knowledge and latent representation $\mut(\bz;\bt)$. %
We consider the external knowledge $\bt$ to be a discrete random variable in a lightly structured, semi-supervised setting. We demonstrate the effectiveness of this approach on multiple  modeling tasks. %
Our contributions are:
\begin{enumerate}[(1),leftmargin=*,itemsep=0pt]
\item We provide a principled, information theoretic approach for injecting side information into neural discrete latent variable models. We provide theoretical backing to show that our methodology captures available information from external knowledge.
\item We define a new model that leads to state-of-the-art results in the semi-supervised setting on two standard event modeling datasets. 
\item We show that our proposed model generalizes the existing state-of-the-art model studied in \citet{rezaee2021event}, where the ``parameter injection'' method developed there can be understood as a special case of our framework.
\item We experimentally show that our model is more robust when the external knowledge is noisy and it outperforms other baselines when the external knowledge is partially observed. 
\end{enumerate}
Our code is available at
{{\url{https://github.com/mmrezaee/RevUp}}}.

\section{Background} 
\label{sec:prelim}
\label{sec:relatedWork}
In this section we present the related work and discuss the connections to our approach. %

\paragraph{Mutual Information (MI)}
The mutual information $\mut(x;y)$ measures how much two random variables $x$ and $y$ depend on each other. It is defined as
$\mut(x;y)= \mathbb{E}_{p(x,y)}\log \frac{p(x,y)}{p(x)p(y)},$
where $\mut(x;y)=0$, when $x$ and $y$ are independent. Conditional MI $\mut(x;y\vert z)$ extends MI to measure the conditional dependence between $x$ and $y$ given $z$, as
$\mut(x;y\vert z)= \mathbb{E}_{p(x,y,z)}\log \frac{p(x,y|z)}{p(x|z)p(y|z)}.$
When $\mut(x;y\vert z)=0$, there isn't any information between $x$ and $y$ that is not present in $z$.

\paragraph{Information Bottleneck Principle (IB)}
The Information Bottleneck principle is a method for finding the most informative encoding of an input $x$ with respect to the target output $y$. This is accomplished by finding the maximally compressed encoding $z$ of $x$ that is most informative about $y$ \citep{tishby99information}. The objective is to maximize 
\begin{align}
\mut(z;y)-\beta\mut(x;z),
\end{align}
where $\mut(z;y)$ denotes $z$ and $y$ mutual information, and $\mut(x;z)$ denotes $x$ and $z$ mutual information. $\beta$ balances compression and informativeness.

\paragraph{Variational Autoencoder (VAE)}
The Variational Autoencoder is similar to the Information Bottleneck Principle in that it finds an encoding $z$ of $x$ to reconstruct $x$. However, the approach is different as the aim is to approximate the intractable posterior distribution $p_{\theta}(z\vert x)$ with a variational distribution $q_{\phi}(z\vert x)$ in order to optimize the Evidence Lower Bound (ELBO) \citep{kingma2013auto}. 


\paragraph{Incorporation of Side Knowledge}
Conceptually, we consider an ``event'' to be a condensed form of knowledge that outlines part of a particular situation. 
Semantic frames are a prime example of relevant side knowledge: a semantic frame can be thought of as an abstraction over highly related events. 
Though they also provide abstractions over the \textit{who}s, \textit{what}s, \textit{where}s and \textit{how}s of events, in this setting, it is sufficient to consider an event's semantic frame as a type of label or cluster id. %
Multiple potential sources of semantic frames exists, e.g., FrameNet, PropBank, or VerbNet. 

Recently, incorporating external knowledge has been investigated by numerous studies in a wide range of tasks beyond NLP~\citep{kang2017incorporating,NIPS2017_0bed45bd,zhang2021cross}, and zero-shot classification~\citep{badirli2021fine}. 
Common across these efforts is treating the side information as part of the input to be encoded. %
This makes the side information prerequisite knowledge for the model to be learned, rather than supplementary. %

A recent approach, called Sequential, Semi-supervised Discrete Variational Autoencoder~\citep[SSDVAE]{rezaee2021event}, is a new method for structured semi-supervised modeling that allows, but does not require, side information to guide the learning in an approach called ``parameter injection.'' 
Because the SSDVAE framework is a deep latent variable model that is specifically designed to treat external knowledge as supplementary to the main task, we focus our study within it and the associated NLP-based computational event modeling tasks examined. %
They define parameter injection as
\begin{definition}
 (Parameter injection) 
 Let $\mathbf{t}\sim \text{GumbelSoftmax}(\mathbf{t};\gamma)$, where $\boldsymbol{\gamma}$ are the logits. 
 If $\mathbf{t}$ is observed as external knowledge and represented with a one-hot vector $\mathbbm{1}(\mathbf{t})$ with $t_{k^*}=1$, 
 the operation $\boldsymbol{\gamma}=\boldsymbol{\gamma}+(\norm{\boldsymbol{\gamma}}*\mathbbm{1}(\mathbf{t}))$ guides the latent variable $\mathbf{t}$ during the training because 
 on average increases the $t_{k^*}$ value
 \citep{maddison2016concrete}.
 \end{definition}
In RevUp, we build on and generalize this notion of parameter injection, in part by introducing the empirical distribution $p_{\mathcal{D}}(\bt\vert\bx)$ to accommodate the dependence between data $\bx$ and knowledge $\bt$. %

\section{RevUp: Revise and Update} 
\label{sec:method}

Previous work \citep[SSDVAE]{rezaee2021event} has empirically shown that incorporating external knowledge in discrete GumbelSoftmax parameters improves model performance with different metrics such as classification, event modeling and training speed. We seek to re-frame this view with information theory terms and generalize it. %
For fairness and consistency we stick with the same types of computational event modeling tasks that SSDVAE was developed for. %
We first describe the probabilistic model we develop (\cref{sec:probEnc}) and the loss/training methodology (\cref{sec:trainingObj}). These enable smoothly incorporating side knowledge into a probabilistic neural model. %
Recall we provide a conceptual overview of RevUp in \cref{fig:RevUpSteps}. %


\subsection{Setup} 
\label{sec:method:eventmodeling}

We assume that within a document, we have a sequence of $M$ different predicate/argument-style events, $\bx = x_1 ... x_M$, with each $x_m$ describing some action (the predicate) occurring among participants of that action (the arguments). %
If each event $x_m$ could also be paired with a semantic frame $t_m$, so we have a corresponding sequence $\bt$, then considered across multiple events, $\bt$ could provide a sequential generalization. %
E.g., in the example event sequence from \cref{fig:eventArch} that includes ``convicted man of murdering,'' and ``sentenced man to death,'' if each event can be associated with a semantic frame (such as \textsc{verdict} and \textsc{sentencing} for these two events), then the corresponding sequence of frames provides both an abstraction over the entire event sequence, and an incredibly rich, yet low-dimensional, collection of side knowledge. %

Having paired sequences $\bx$ and $\bt$ is not restrictive. Whether $\bt$ is partially observed (i.e., not all $x_m$ have observed frames $t_m$) or noisy (i.e., the observed $t_m$ for $x_m$ is incorrect) during training, our approach can still extract useful signal. %
 
\subsection{Probabilistic Encoding}
\label{sec:probEnc}
Our problem setup is that we have input text $\bx$ describing some complex situation, paired with a partially observed sequence of frames $\bt$. %
To account for when side knowledge is/isn't observed, we define a knowledge indicator set $\boldsymbol{l}=l_m\vert_{m=1}^{M}$ with $l_{m}\in\{0,1\}$, where $l_{m}=1$ denotes the external knowledge $t_m$ is present and $l_{m}=0$ means the external knowledge is not available (latent).  %
To enable successful modeling, we introduce $\bz=z_m\vert_{m=1}^M$, a set of $M$ latent variables to first compress the information of the given inputs $\bx$ and then be informative regarding $\bt$. We define $p_{\theta}(\bz\vert\bx)$, a probabilistic encoder from data points $\bx$ to the latent variables $\bz$, parameterized by a neural network $\theta$.

We define a joint model over $\bt, \bx$ and $\bz$ as
$p(\bt,\bx,\bz;\boldsymbol{l})
  =
  p_{\mathcal{D}}(\bx)
  \prod_{m} p_{\theta}(z_m\vert t_{m-1},\bx)
  p_{\gamma}
  \big(
    t_m\vert x_m, z_{m};l_m
  \big),
$
where $p_{\mathcal{D}}(\bx)$ is the empirical distribution over the input variables $\bx$. Consistent with previous approaches, 
$p_{\theta}(z_m\vert t^{(s)},\bx;l_m)$ is a Gaussian. %
Similarly
$
  p_{\gamma}
  \big(
    t_m\vert x_m, z_m;l_m
  \big)
$
posits a distribution over semi-supervised latent knowledge $\bt$ given $\bx$ and latent variables $\bz$. %
To learn richer representations, we force  $\bz$ and $\bt$ to depend on one other in a sawtooth fashion: $t_i$ depends on $z_i$, but $z_i$ depends on $t_{i-1}$. %
\Cref{fig:eventArch} shows an illustration. %
This is a novel segmented, autoregressive sequencing for event modeling. 


We define $r_{\gamma}(\bt\vert\bz)$ as the \textit{proposed distribution} that relates $\bz$ to $\bt$, where $\gamma$ is computed as the outcome of a neural encoding of $\bz$, $\text{NN}(\bz)$. This distribution must be learned.
For simplicity we omit the node index $m$ when possible. %

\paragraph{Revision Phase}
Incorporating the external knowledge in the training phase must satisfy two criteria: First, without observation $(l_m=0)$, we just rely on $r_{\gamma}(t_m|z_m)$. Second, when we have access to external knowledge, $p_{\mathcal{D}}(t_m\vert x_m)$ should be used for \textit{guidance} and not discarding the proposed distribution.
We define the \textit{revised distribution} 
$p_{\gamma}
  \big(
    t_m\vert x_m,z_m;l_m
  \big)$
as
a smoothed weighted average of the proposed distribution 
and the empirical distribution as 
$
p_{\gamma}
  \big(
    t_m\vert x_m,z_m;l_m
  \big)
  =
  \dfrac
  {1}{1+\lambda l_m}
  r_{\gamma}(\st_m\vert z_m)
  +
\dfrac{
  \lambda
  l_{m}
  }
  {
  1
  +
  \lambda
  l_{m}
  }
p_{\mathcal{D}}{(\st_m\vert x_m)},
$
where $\lambda \in \mathbb{R}^+$ is a weighting parameter for balancing the proposed and empirical distributions. %
In this setting, $\lambda$ depends on the level of confidence in our external knowledge, where for less noisy knowledge, we can choose higher values for $\lambda$.
In practice, we found that $\lambda=1.0$ works reasonably.\footnote{In dev experiments we tried various values of $\lambda$, both large and small. When $\lambda$ was high (greater weight to the empirical distribution), the model over-emphasized this signal, which was detrimental at test time when we intentionally withheld this signal. In contrast, when $\lambda$ is low (greater weight to the proposed distribution), there is less emphasis on the empirical distribution. However, as that empirical distribution could itself be quite sparse (due to $\epsilon$), the model learns to disregard it. This highlights a strength in simplicity of the equal weighting.} %

If side knowledge is absent ($l_m=0$), we have  $\hat{\lambda}_m=1.0$ so $p_{\gamma}(t_m\vert x_m,z_m)=r_{\gamma}(t_m\vert z_m)$: the model just uses the proposed distribution. This allows gradients to propagate through the network. %
During the test phase we do not use any external knowledge, so the revised distribution $ p_{\gamma}
  \big(
    \bt\vert\bx,\bz;\boldsymbol{l}
  \big)$
  reduces to the proposed distribution $r_{\gamma}(\bt\vert\bz)$. 

{\textbf{\emph{Analysis and Insights}}}
If we define $\hat{\lambda}_{m}={1}/{(1+\lambda l_m)}$, we rewrite this as
$
 p_{\gamma}
  \big(
    t_m\vert x_m,z_m;l_m
  \big)
  = 
  \hat{\lambda}_m
    r_{\gamma}(t_m\vert z_m)
    + 
  (1-\hat{\lambda}_m)
  p_{\mathcal{D}}{(\st_m\vert x_m)}.
$
This slight notation change is beneficial, as it lets us characterize the behavior of our revision step in terms of expected observability of side knowledge. %
Specifically, in a semi-supervised setting, if the probability of observing any particular piece of side information can be modeled as $l \sim \text{Bern}(\epsilon)$, where $\epsilon$ is the observation probability, by marginalizing out $l$ we have 
$\mathbb{E}_{l\sim \mathrm{Bern}(\epsilon)}
p_{\gamma}(t|x,z,l)=
  \hat{\epsilon}
  r_{\gamma}(t\vert z)
  +
  (1-\hat{\epsilon})
  p_{\mathcal{D}}{(t\vert x)},
$
where $\hat{\epsilon}=1-\left(\frac{\lambda}{1+\lambda}\right)\epsilon.$
This indicates that with more observation, we rely more heavily on the empirical distribution and less on the proposed distribution. %
For space, see \Cref{app:dropOut} for more details/the derivation. %


Finally, our framework generalizes SSDVAE's parameter injection (see \Cref{app:ssdvaeGenProof}):

\begin{thm}
If $r_{\gamma}(\st\vert z)=\text{Cat}(\boldsymbol\gamma)$  , a categorical distribution with parameters $\gamma$, and the empirical distribution
$p_{\mathcal{D}}{(t\vert x)}$ is a one-hot representation with $t_{k^*}=1$, the revision step reduces to SSDVAE parameter injection.
\label{thmGeneralize}
\end{thm}

\noindent We have described how to guide the latent variable $t$ in the encoding phase. %
Next we define our objective function to capture the background information and decoding $\bt$ to effectively model event descriptions.

\subsection{Training Objective and Decoding}
\label{sec:trainingObj}
After training, 
the model relies solely on the proposed distribution $r_{\gamma}(t_m|z_m)$ to predict $t_m$, implying the statistics of $t_m$ will only depend on $z_m$. %
To capture the background information in $\bt$, 
minimize $\mut(\bx;\bt\vert\bz)$ during training.
In information theory terms, in the ideal situation where $\mut(\bx;\bt\vert\bz)=0$, there is not any residual information between $\bx$ and $\bt$ that is not captured by the latent representation $\bz$ \citep{kirsch2020unpacking}. Therefore without the need of $\bx$ and $p_{\mathcal{D}}(\bt\vert\bx)$, the latent $\bz$ is enough to predict $\bt$. %

To be meaningfully learned, $\bt$ must be informative enough to make some prediction or reconstruction. For clarity, we refer to the targets as $\by$, though for a task like language modeling, $\bx = \by$. 
To achieve this learning, we maximize $\mut(\by,\bt)$.

Together, our ideal objective is
$
  \mathcal{L}
  =
  -\mut(\by;\bt)
  +\alpha \mut(\bx;\bt\vert\bz),
  $
where $\alpha$ is a tunable hyperparameter. %
This is difficult to optimize because within each $\mut$ term there are intractable marginalizations (such as over $\bx$). %
To understand why this is our ideal objective, we show that maximizing the intractable mutual information $\mut(\bt;\bz)$ is inherently included in this unified objective for the reconstruction tasks: %

\begin{thm}
For tasks where we maximize $\mut(\bx;\bt)$: minimizing $\mut(\bx;\bt\vert\bz)$ leads to maximizing $\mut(\bt;\bz)$.  
\label{theoremMut}
\end{thm}

\noindent This theorem shows that reconstruction tasks like language modeling explicitly maximize the mutual information between different data representations, which is consistent with the InfoMax principle. 
See \cref{app:mutalInfo} for the proof. As a consequence of \Cref{theoremMut}, we are maximizing the mutual information between two views of $\bx$: compressed representation $\bz$ and side information $\bt$. 

With that understanding, we proceed to a tractable approximation for our objective. Following \citet{alemi2016deep}, we have
$
-\mut(\by;\bt)
  =
  -\mathbb{E}_{p(\by,\bz,\bt)}
  \log{
  \frac{
  p(\by\vert\bt)}{p(\by)
  }
  }.
$
We approximate this as
\begin{align}  
-\mut(\by;\bt)
  & \leq
  -\mathbb{E}_{p(\by,\bz,\bt)}
  \log 
  \dfrac
  {q_{\phi}(\by\vert\bt)}
  {p(\by)} 
  \label{eq:mut_y_t_full}
  \\
  & = 
  \underbrace{
  -\mathbb{E}_{p(\by,\bz,\bt)}
  \log 
  {q_{\phi}(\by\vert\bt)}
  }_{\mathcal{L}_{y}}
 - H(\by).
  \label{eq:taskRecons}
\end{align}
Since $p(\by\vert\bt)$ is intractable, we approximate it via a decoder $q_{\phi}(\by\vert\bt)$ that can be computed by a neural network, denoted as $\phi$.  As the task entropy $H(\by)$ is constant, we just minimize $\mathcal{L}_y$. %
While not explicitly reflected in \Cref{eq:mut_y_t_full}, note that  irrespective of whether side information is present or not, $\mut(\mathbf{y};\mathbf{t})$ depends on 
$
r_{\gamma}(\mathbf{t}\vert \mathbf{z})
$. 
See \Cref{sec:app:I_y_t} for additional analysis of \cref{eq:taskRecons}.

\paragraph{Updating Phase}
The term
$
  \mut(\bx;\bt\vert\bz) 
  =
  \mathbb{E}_{p_{\gamma}(\bt,\bx,\bz)}\log\frac{
  p_{\gamma}(\bt\vert\bx,\bz;\boldsymbol{l})}{
  p(\bt\vert\bz)}
$
makes $\bz$ informative about $\bt$. %
We approximate with a surrogate objective $\mathcal{L}_{\mut}$:
\begin{align}
  \mut(\bx;\bt\vert\bz)  
  &\leq
\mathbb{E}_{p(\bt,\bx,\bz)}\log \dfrac {p_{\gamma}(\bt\vert\bx,\bz)}{r_{\gamma}(\bt\vert\bz)}
=\mathcal{L}_{\mut}
.
\label{eq:condMutApprox}
\end{align}

\noindent
This surrogate objective encourages the proposed distribution $r_{\gamma}(\bt\vert\bz)$ to be updated to be close to the revised distribution $p_{\gamma}(\bt\vert\bx,\bz;\boldsymbol{l})$. %
After training, the proposed distribution $r_{\gamma}(\bt\vert\bz)$ plays the role of $p_{\gamma}(\bt\vert\bz,\bx)$ and does not need to explicitly use $p_{\mathcal{D}}{(\bt\vert\bx)}$. %
Throughout, we refer to $\mathcal{L}_{\mut}$ as \textit{updating}. 
The updating objective for our model is given by
\begin{align}
\mathcal{L}_{\mut}=\sum_{m=1}^M 
\infdiv
{p_{\gamma}(t_m \vert x_m, z_m; l_m)}{r_{\gamma}(t_m\vert z_m)},
\end{align}
where by expanding further we obtain
\begin{align}
\mathcal{L}_{\mut}
& = 
 \sum_{m=1}^{M}
 \biggl[-H(t_m\vert x_m,z_m)
+ \hat{\lambda}_m H(t_m\vert z_m) 
\nonumber
\\
& - (1-\hat{\lambda}_m) 
\mathbb{E}_{p_{\mathcal D}(t_m|x_m)}
\log r_{\gamma}(t_m\vert z_m)\biggr].    
\end{align}

\noindent This sum is computed over observed nodes. For unsupervised nodes, the revised distribution and proposed distribution are equal and their KL terms are zero.
The last term is a classification term. 

\paragraph{Regularization}
To improve the model generalization ability, we introduce regularization terms for $\bz$ and $\bt$.
We constrain the mutual information between data $\bx$ and $\bz$ latent representations, 
$
  \mut(\bx;\bz)  
  =
  \mathbb{E}_{p(\bz,\bx)}\log\frac{p_{\theta}(\bz\vert\bx)}{p(\bz)}
$ as
\begin{align}
  \mut(\bx;\bz)  
  &
  \leq
  \underbrace{
  {\infdiv{p_{\theta}(\bz\vert\bx)}{q(\bz)}}
  }_{\mathcal{L}_z},
  \label{eq:z-regularization}
\end{align}
where we introduce a variational distribution $q(\bz)$ due to intractability of $p(\bz)$.
For simplicity we assume that $q(\bz)$ factorizes over independent Gaussian random variables
as
$
q(\bz)
=\prod_{m=1}^{M}
q(z_m)$
, where 
the variational distribution over $z_m$
is given by the unit Gaussian
$
z_m\sim \mathcal{N}{(\mathbf{0},\mathbf{I})}.
$ 
Here $\mathcal{L}_{z}$ is estimated with
the standard Monte-Carlo sampling:
\begin{align}
 \mathcal{L}_z\approx
       \dfrac{1}{S}
  \sum_{m=1}^{M}
  \sum_{s=1}^{S}
  \infdiv{p_{\theta}(z_m\vert t^{(s)}_{m-1},\bx)}{q(z_m)}.
\end{align}
We reduce the distance between the proposed distribution $r_{\gamma}(\bt\vert\bz)$ and a uniform distribution $\mathcal{U}(\mathbf{t})$:
\begin{align}
\mathcal{L}_t=
    \mathbb{E}_{p_{\mathcal{D}}(\bx)p_{\theta}(\bz\vert\bx)r_{\gamma}(\mathbf{t}\vert\bz)} 
    \log \dfrac{r_{\gamma}(\mathbf{t}\vert\bz)}{\mathcal{U}(\mathbf{t})}.
\label{eq:t-regularization}
\end{align}
The Kullback-Leibler (KL) divergence terms in \Cref{eq:z-regularization} and \Cref{eq:t-regularization} help avoid overfitting. An alternative interpretation for these regularization terms is discarding the task-irrelevant information. %
We combine \Cref{eq:taskRecons,,eq:condMutApprox,,eq:z-regularization,,eq:t-regularization} to at our final objective
\begin{align}
    \mathcal{L}^{U}
    =
    \mathcal{L}_y
    +\alpha\mathcal{L}_\mut
    +\beta\mathcal{L}_z
    +\zeta\mathcal{L}_t,
    \label{eq:overall-objective}
\end{align}
where $\beta$ and $\zeta$ denote the trade-off parameters, and can be set empirically, as described in \cref{app:setup}.

\begin{figure*}
    \centering
    \begin{subfigure}[t]{0.32 \textwidth}
    \includegraphics[width=1.0\linewidth]{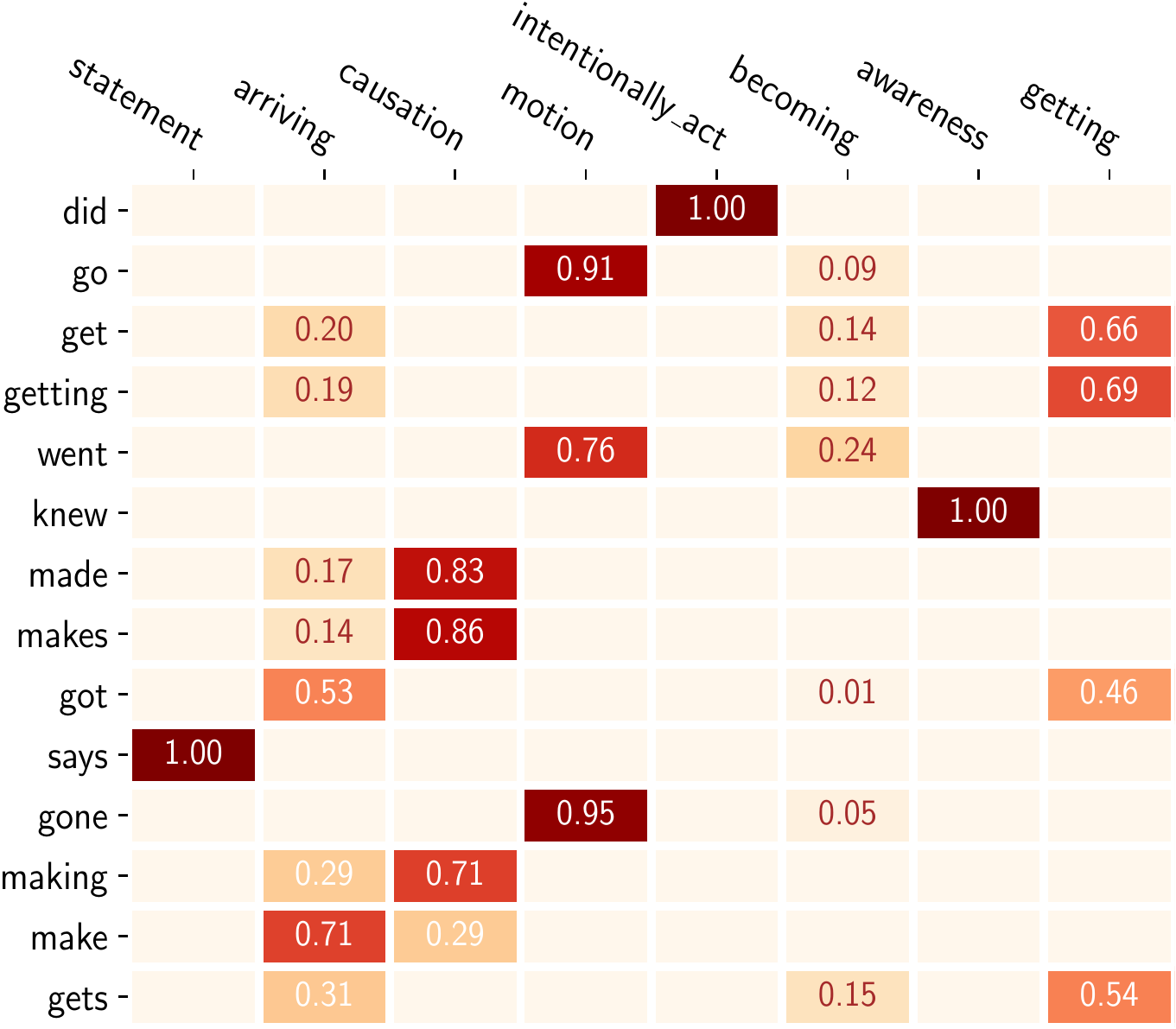}
        \caption{Normalized samples from learned $r_{\gamma}(t|z)$ given predicates. %
        }
        \label{fig:toy-iters:approx}
    \end{subfigure}
~
\begin{subfigure}[t]{0.32\textwidth}
    \includegraphics[width=1.0\textwidth]{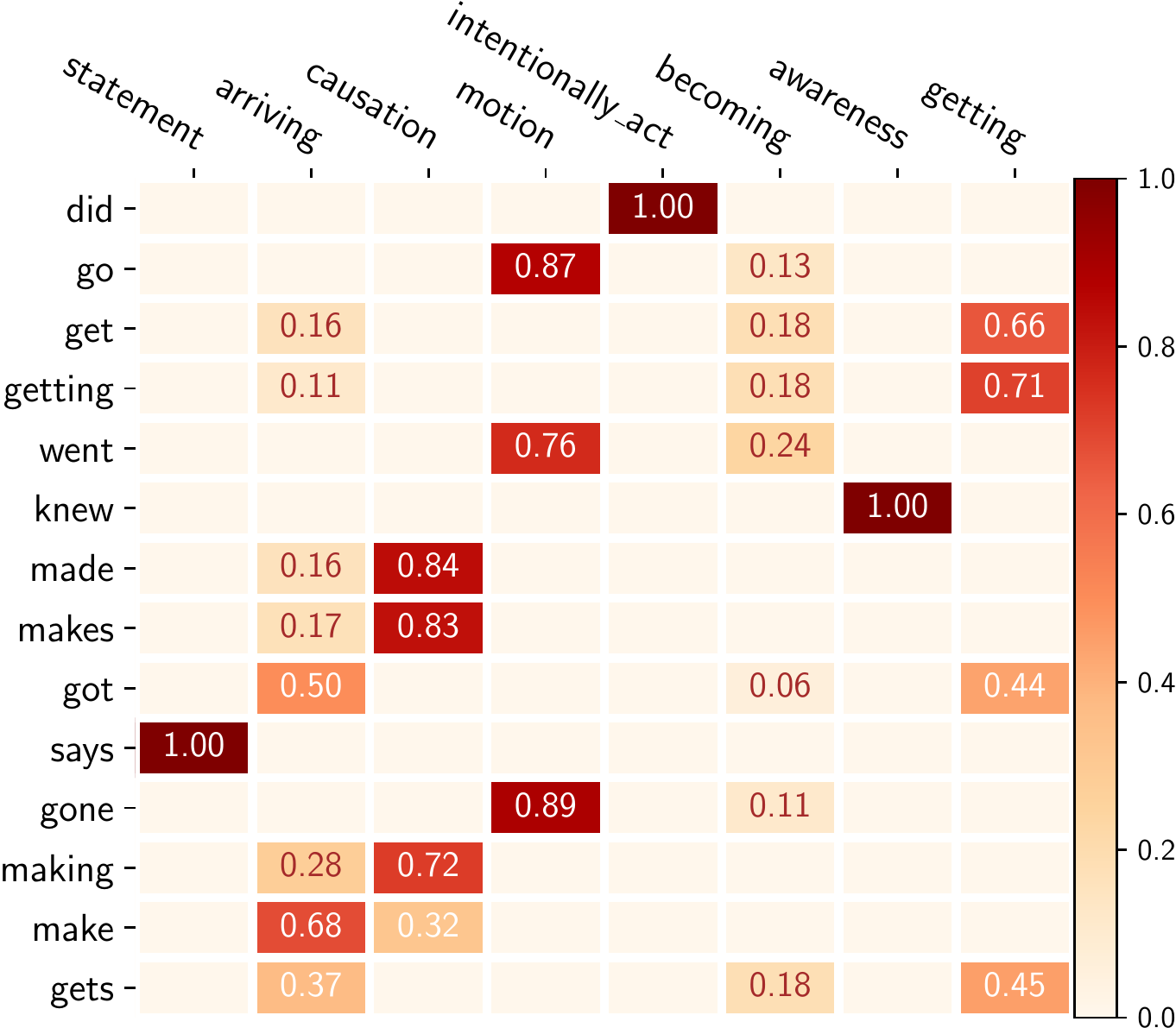}
    \caption{Ground truth conditional distribution of semantic frames given predicates. %
    }
    \label{fig:toy-iters:gold}
\end{subfigure}
~
    \begin{subfigure}[t]{0.327\textwidth}
        \includegraphics[width=1.0\textwidth]{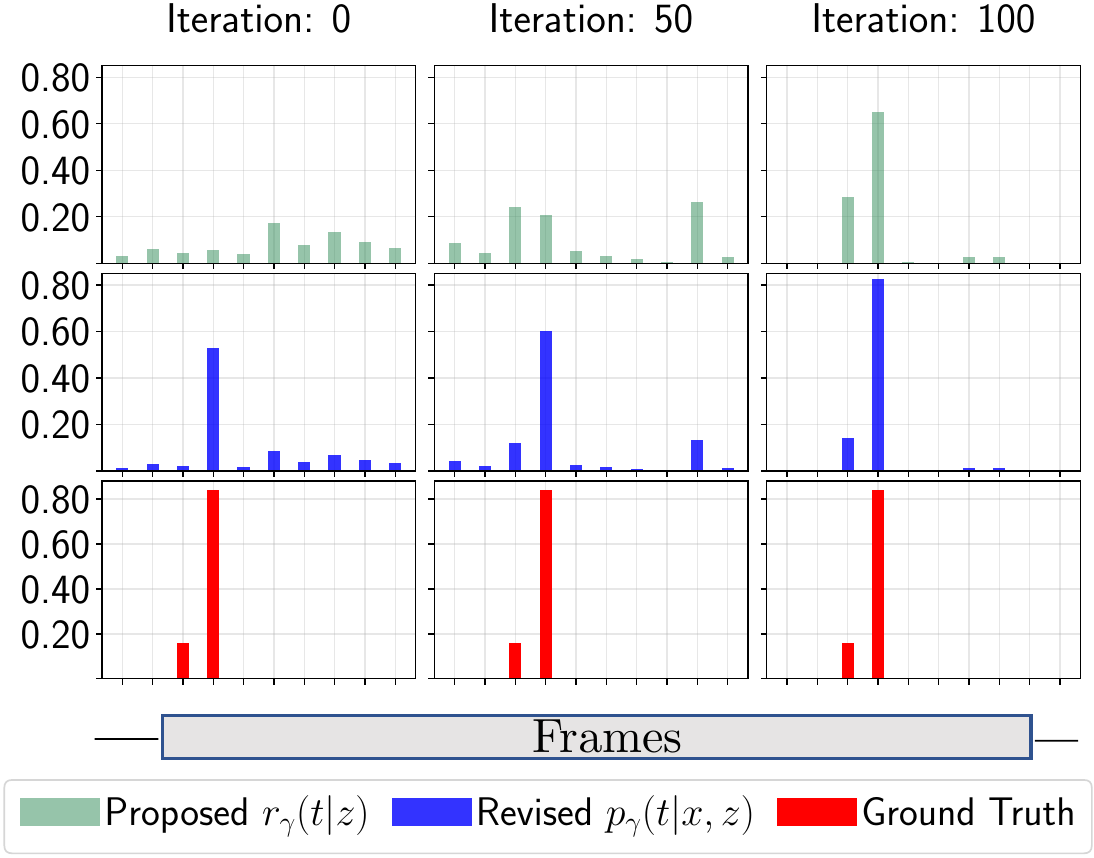}
        \caption{The proposed and revised distributions converge to the ground truth distribution.
        }
        \label{fig:toy-iters:iters}
    \end{subfigure}
    \caption{A demonstration of RevUp working on a small dataset with 10 frames. %
    Ground truth frame distributions are shown in \cref{fig:toy-iters:gold}, where in \cref{fig:toy-iters:approx} we show that the distribution $r_\gamma(t|z)$ that RevUp learns very closely matches the ground truth. %
    In \cref{fig:toy-iters:iters}, we show how RevUp affects both the proposed and revised distributions. %
    Initially, the $r_{\gamma}(t|z)$ prediction is random. After 50 iterations it is closer to the revised distribution $p_{\gamma}(t|z,x)$. After 100 iterations of training, both of them get close to the ground truth distribution. %
   }
   \label{fig:toy-iters}
   \vspace{-2ex}
   \end{figure*}

\subsection{Architecture For Event Modeling}
\label{sec:arch-event-modeling}
To ensure fair comparisons,
we focus on the reconstruction task similar to a $\beta$-VAE framework \citep{higgins2016beta}. The overall structure is depicted in 
\Cref{fig:eventArch}.
Following previous work on event modeling \citep{pichotta2016learning,rezaee2021event,weber2018hierarchical,gao-etal-2022-improving}, we represent each document $\bx$ as a sequence of $M$ events, where each event is a 4 tuple of predicate (verb), two main arguments (subject and object), and a modifier (if applicable). Each event is associated with a discrete semantic frame. E.g., \textit{convicted man murdering of} is an event and the semantic frame is \fsc{[{Verdict}]}.
All the possible frames are collected in a vocabulary of size $T$. 
In this setting, we obtain a point estimate for $p_{\mathcal{D}}(t_m|x_m)$ as $\delta(x_m,t_m)$. Sampling from this empirical distribution outputs a one-hot vector of dimension $T$. The proposed distribution $r_{\gamma}(t|z)$ is a Gumbel-Softmax distribution. %
We found that the Gumbel-Softmax algorithm is particularly suitable for our task because it can effectively approximate discrete distributions and backpropagate gradients. While experiments with the Straight-Through Gumbel-Softmax (STGT) yielded near identical performance to the Gumbel-Softmax method, we opted for the latter. STGT generates one-hot vectors during the forward pass, but requires an approximation of the gradient using Gumbel-Softmax samples. By setting a low temperature of 0.5, the generated Gumbel-Softmax samples become almost identical to one-hot vectors, eliminating the need for gradient approximation. The effects on the gradient were carefully analyzed and the results are presented in \citet{damadi2022gradient}, where a comprehensive study of gradient properties was conducted.
To learn richer representations, we define an embedding matrix $\mathbf{E}\in\mathbb{R}^{T\times d_t}$, to convert a simplex frame into a vector representation as $e_m=t_m^{T}\mathbf{E}$. 

%

\paragraph{Encoding and Decoding}
With data point $\bx\sim p_{\mathcal{D}}(\bx)$, we encode the whole sequence into recurrent hidden representations $\mathbf{H}=\{h\}_{t=1}^{T}$. For each event $m$, we draw Gaussian random variable $z_m \sim
\mathcal{N}{(\mu_m,\sigma_m)}
$
where $\mu_m$ and $\sigma_m$
are the outputs of attention layers over frame embedding $e_{m-1}$ and hidden states $\mathbf{H}$. 
We use a linear mapping over $z_m$ to compute Gumbel softmax parameters for the proposed distribution $r_{\gamma}(t_m\vert z_m)$. %
Given the proposed distribution $r_{\gamma}(t_m\vert z_m)$ and the empirical distribution $p_{\mathcal{D}}(t_m\vert x_m)$, we first draw a Bernoulli sample 
  $
 \hat\lambda_m 
 $, then we draw knowledge sample from the mixture of probabilities by
$
t_m \sim
p_{\gamma}(t_m\vert x_m,z_m;l_m).
  $
  From the encoding phase,
  all the embedding vectors are gathered into $\mathbf{e}=\{e_m\}_{m=1}^{M}$.  
  At generation time, analogously to the encoder, we use an attention layer over the decoder recurrent hidden state $h$ and frame embeddings $\mathbf{e}$, resulting in decoder logits $g$. The generative distribution over possible next tokens is given by $x_t \sim p(x_t|g)$.

\begin{figure*}
    \centering
    \begin{subfigure}[b]{0.242\textwidth}
    \includegraphics[width=1.0\linewidth]{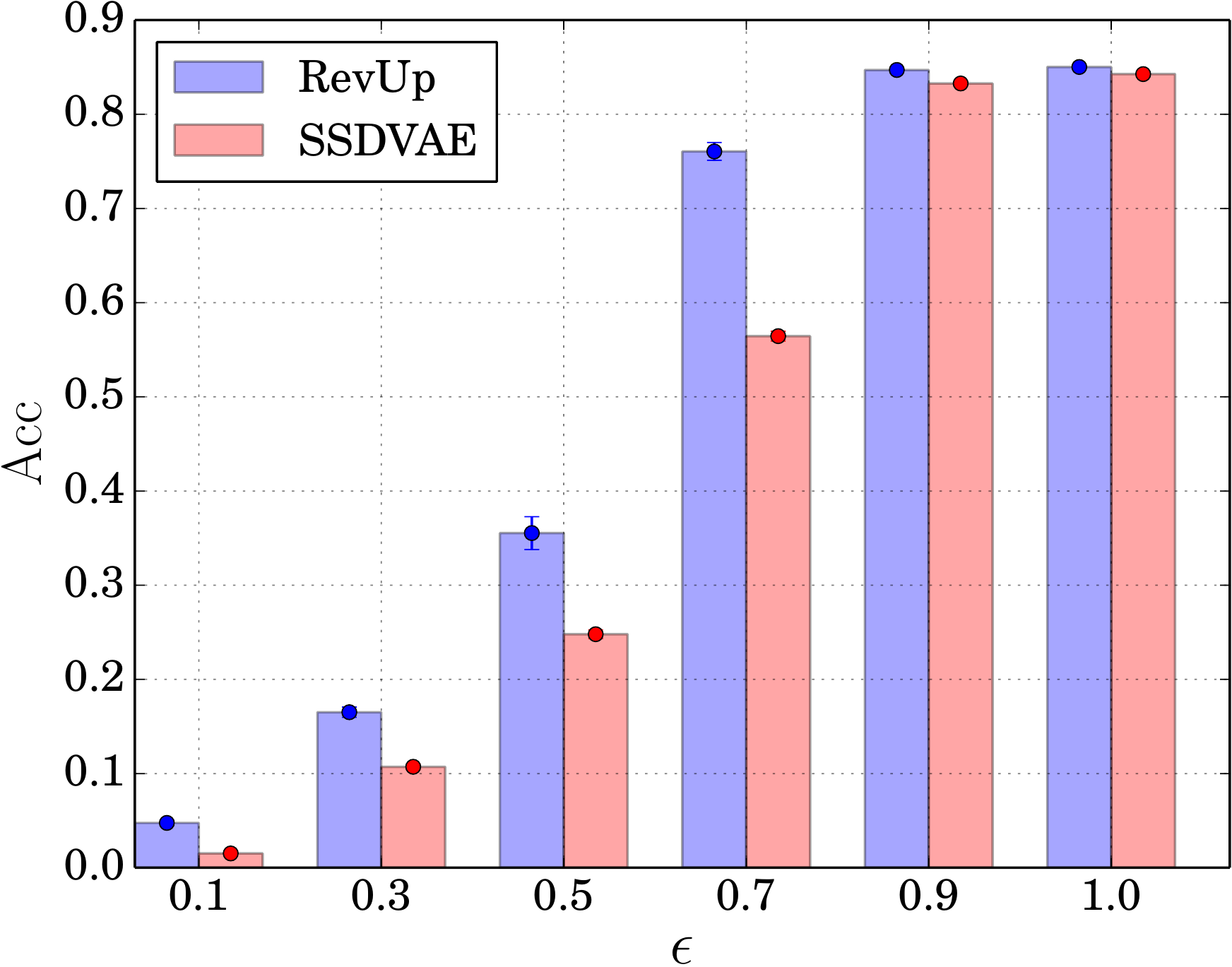}
        \caption{NYT Accuracy}
        \label{fig:accNyt}
    \end{subfigure}
  \vspace{0.1em}
\begin{subfigure}[b]{0.242\textwidth}
    \includegraphics[width=1.0\textwidth]{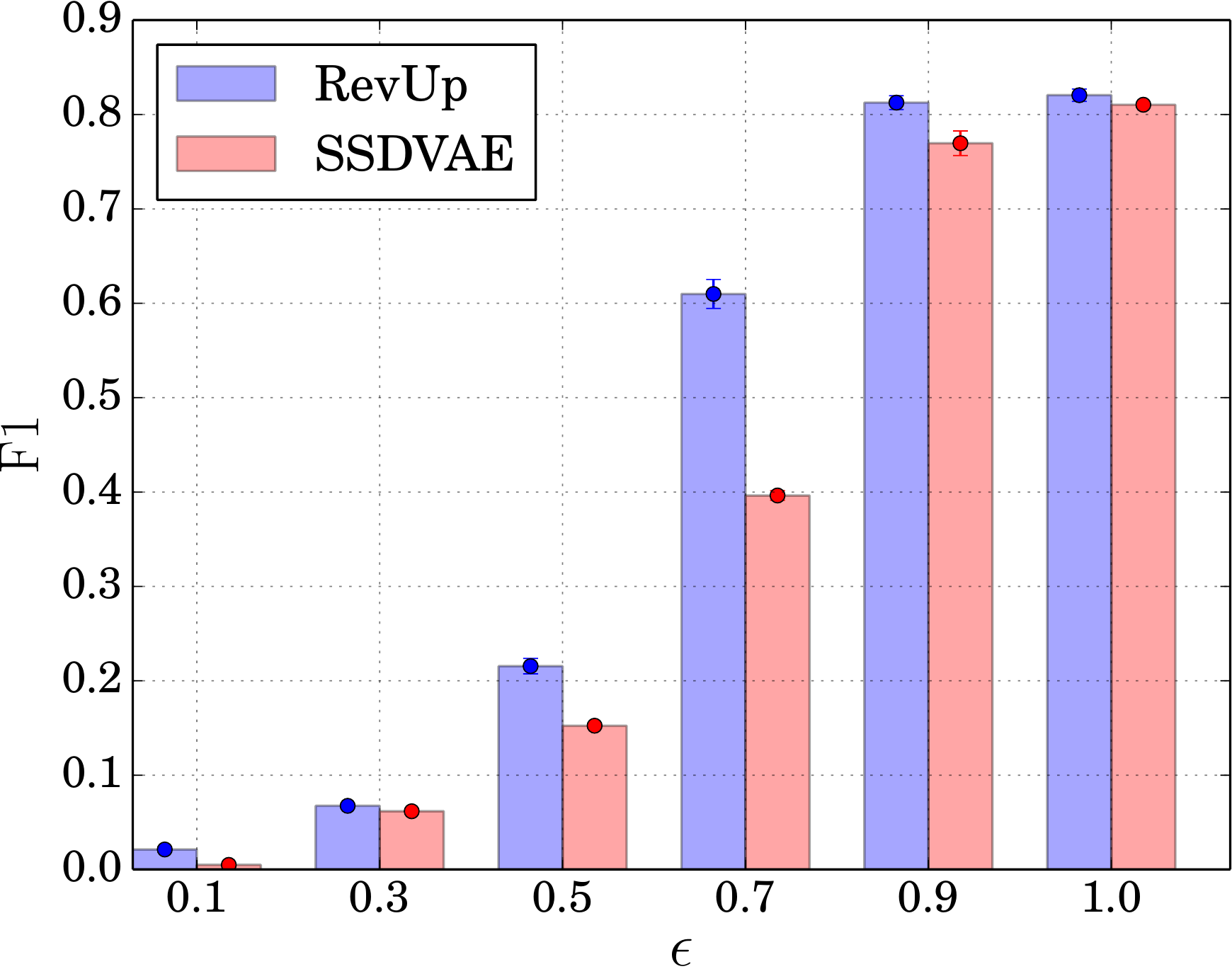}
    \caption{NYT F1 score}
    \label{fig:f1NYT}
\end{subfigure}
~
    \begin{subfigure}[b]{0.242\textwidth}
        \includegraphics[width=1.0\textwidth]{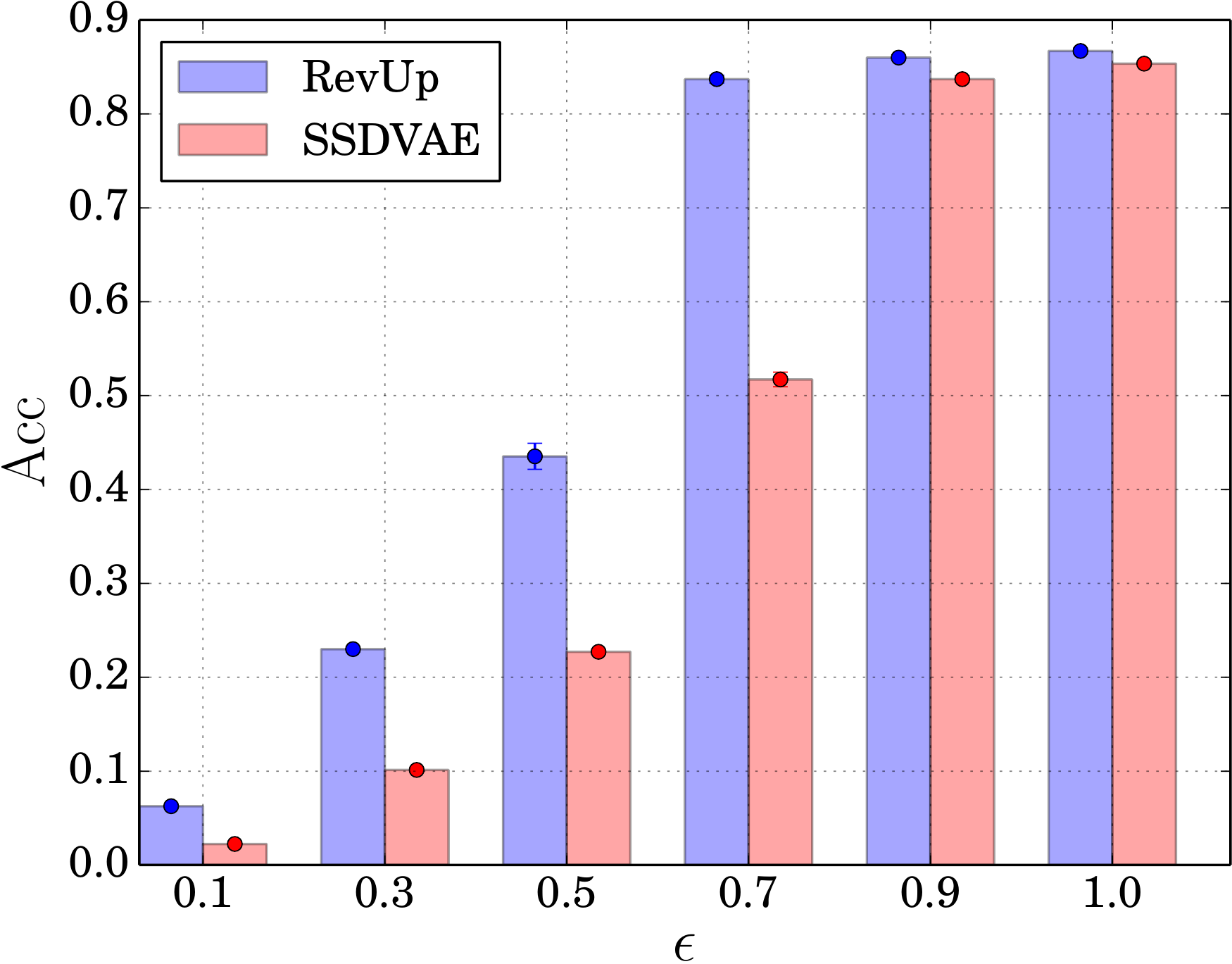}
        \caption{Wikipedia Accuracy}
        \label{fig:accWiki}
    \end{subfigure}
    \begin{subfigure}[b]{0.242\textwidth}
        \includegraphics[width=1.0\textwidth]{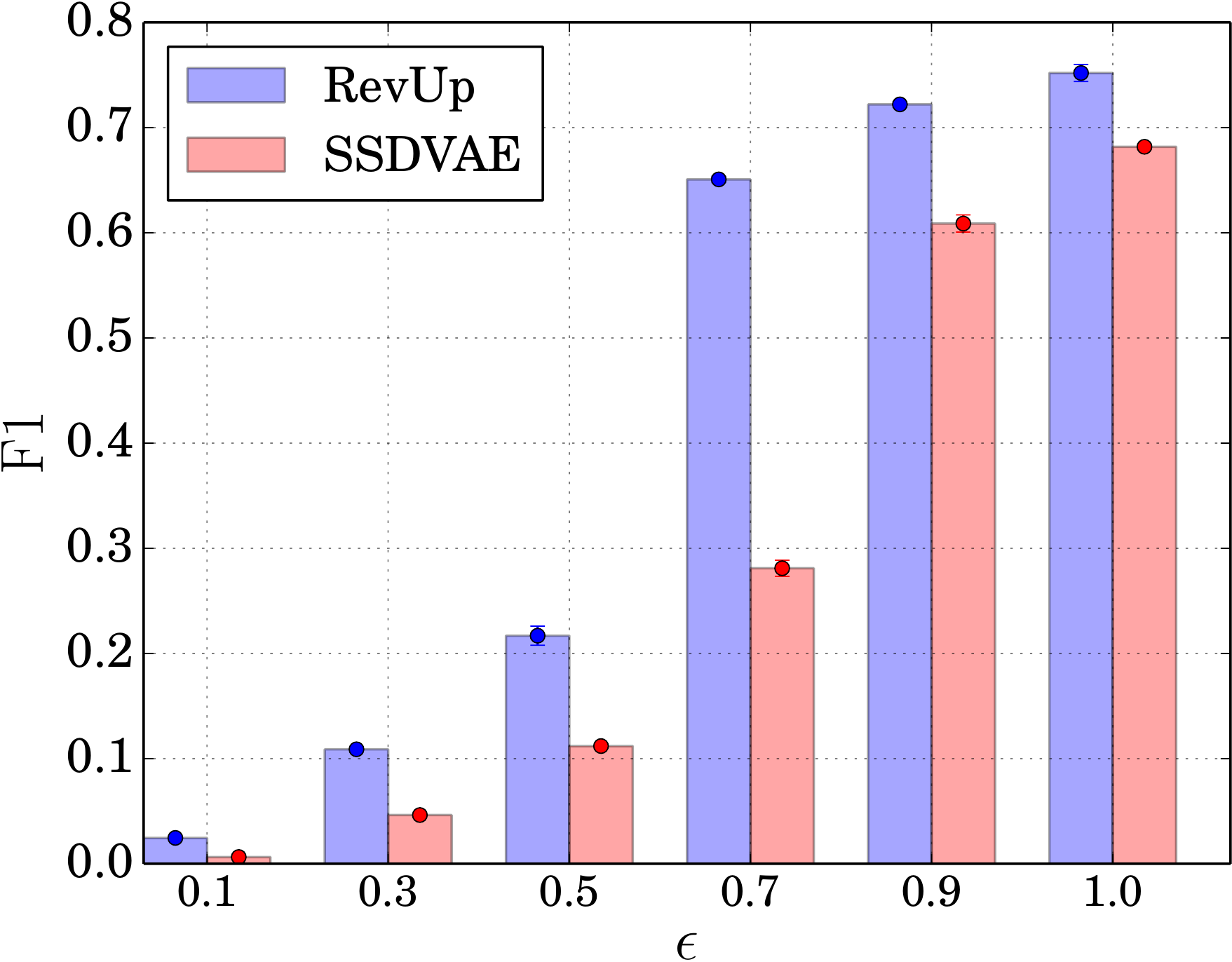}
        \caption{Wikipedia F1 score}
        \label{fig:f1Wiki}
    \end{subfigure}
   \caption{RevUp vs SSDVAE accuracy for sequential semantic frame classifications for NYT and Wikipedia dataset. See \Cref{app:addExperiments} for full results. For all degree of observation probabilities, RevUp outperforms the baseline.
   }
   \label{fig:Classification}
   \vspace{-2ex}
   \end{figure*}
\section{Experimental Results} 
\label{sec:experiments}
Following previous work \citep{rezaee2021event}, we experiment on event modeling tasks using Concretely Annotated versions of New York Times Gigaword (NYT) and Wikipedia datasets \citep{ferraro2014concretely}. Both are English and have FrameNet semantic frames provided via SemaFor~\citep{das2014frame}. %
We perform a direct comparison with the previous state-of-the-art work \citep{rezaee2021event} by using the same frame types, which were derived from the FrameNet annotations on Wikipedia articles. %
Training, validation and test splits for NYT and Wikipedia have 320k/17k/7k and 240k/8k/11k documents, respectively. For both the validation and test phases we set $l_m=0$ (unsupervised). %
We average results over three runs (standard deviations in the appendix). %
We tuned the hyperparameters on the validation dataset. %
See \Cref{app:setup} for additional implementation and data details.


\subsection{Small Dataset Example}
We first examine RevUp by examining its behavior on a small, focused example dataset. %
We sampled 400 newswire documents from the NYT dataset. %
We trained a RevUp model, with 10 semantic frame types (10 options for each $t$) and where each $z$ was 100 dimensional. %
To obtain the ground-truth distributions of frames given events, we just focus on the predicates and collect all the semantic frames given that specific predicate. We carefully selected the frame types and data to reflect a diverse range of difficulties.
For example, events including the \textit{did} predicate are always associated with the \textsc{intentionally\_act} frame and the corresponding semantic frames for the \textit{gets} predicate are \textsc{arriving}, \textsc{becoming} and \textsc{getting}. %
RevUp predictions are acquired from the revised distribution $r_{\gamma}(t|z)$: given an event $x$ we first draw a Gaussian sample $z$  and then use $\argmax(t)$ to find the proposed index. Finally, we normalize across all the proposed frames. %
We visualize the revision and update steps for the predicate ``\textit{made}'' in \Cref{fig:toy-iters:iters}. 

As evidenced by \Cref{eq:condMutApprox}, the minimization of the KL-divergence between the proposed distribution $r_{\gamma}(t\vert z)$ and the revised distribution $p_{\gamma}(t\vert x,z)$ results in the elimination of non-essential information.
\Cref{fig:toy-iters:iters} gives a visual representation of this phenomenon. At iteration 0, the proposed distribution $p_{\gamma}(t\vert x,z)$ is almost uniform, while the revised distribution $r_{\gamma}(t\vert z)$ is more aligned with the ground distribution. With additional iterations, reaching 100, the gap between these distributions reduces, ultimately leading to convergence with the ground truth for both distributions.

Additionally, our framework is able to effectively capture the conditional distribution of frames given predicates in diverse scenarios, as demonstrated in the heatmap figures in \Cref{fig:toy-iters:approx,fig:toy-iters:gold}. For instance, certain predicates such as ``did,'' ``says,'' and ``knew'' have a single associated frame of ``intentionally\_act,'' ``awareness,'' and ``statement,'' respectively. Meanwhile, the predicate ``made'' has two possible semantic frames of ``arriving'' and ``causation.'' In some cases, such as the predicate ``get,'' there are even three possible frames of ``arriving,'' ``becoming,'' and ``getting.'' The comparison between the ground truth distribution of frames given predicates and the normalized samples highlights the accuracy of our model in capturing these conditional distributions, with minimal error.

\subsection{Baselines}
We compare the proposed RevUp method with the following event modeling approaches.
\begin{enumerate*}[(a)]
    \item \textit{\bf RNNLM}~\citep{pichotta2016learning}: A Bidirectional GRU cell with    two layers, hidden
     dimension of 512, gradient clipping at 5 and Glove 300 embeddings to represent words. We used the implementation provided in \citep{weber2018hierarchical}.
    \item \textit{\bf HAQAE} \citep{weber2018hierarchical}: An unsupervised VAE-based method for script learning and generation. They use vector quantization (VQ-VAE) to define a hierarchical discrete latent space.
    \item\textit{\bf SSDVAE} \citep{rezaee2021event} A semi-supervised VAE-based model. 
    They utilize Gumbel softmax and use the parameter injection to incorporate the side knowledge.
\end{enumerate*}
 The architecture of our model
 largely follows the SSDVAE and HAQAE models with minor modification; see \Cref{app:setup}. 

\subsection{Effect of Noisy Knowledge}
We empirically compare the predictive performance of RevUp and SSDVAE with noisy knowledge.
To do so, in a fully supervised setting, for each event $x_m$, instead of using the associated semantic frame $t_m$, with probability $\eta$ we replace $t_m$ with a random semantic frame $\hat{t}_m$. We train both models with this new training dataset. During the testing phase, we compare the predicted knowledge with ground-truth one. Results in \Cref{tab:effNoise} show that classification and parameter injection in SSDVAE are not enough for capturing knowledge. These results validate the effectiveness of our information capturing strategy when the external knowledge is noisy. 
As we increase $\eta$, SSDVAE performance degrades much more than RevUp.
As an instance when $\eta=0.9$,  SSDVAE's accuracy is almost zero but our proposed model can achieve 0.41.

\begin{table}[t]
\centering
\resizebox{0.8\linewidth}{!}{
\begin{tabular}{l|c|ll|ll}
\hline
\multicolumn{1}{c|}{\multirow{2}{*}{model}}& \multirow{2}{*}{$\eta$}& \multicolumn{2}{c|}{Wiki}       & \multicolumn{2}{c}{NYT}         \\ \cline{3-6}
\multicolumn{1}{c|}{}                    &                      & \multicolumn{1}{l|}{Acc} & F1   & \multicolumn{1}{l|}{Acc} & F1   \\ \hline
SSDVAE                                      &\multirow{2}{*}{0.1}                & \multicolumn{1}{l|}{$0.77$}& {$0.43$} & \multicolumn{1}{l|}{$0.76$}& {$0.55$} \\ \cline{3-6}
RevUp                                       &                  & \multicolumn{1}{l|}{$\mathbf{0.85}$}& {$\mathbf{0.58}$} & \multicolumn{1}{l|}{$\mathbf{0.83}$}& {$\mathbf{0.71}$} \\ \hline
SSDVAE                                      & \multirow{2}{*}{0.2}                   & \multicolumn{1}{l|}{$0.69$}& {$0.35$} & \multicolumn{1}{l|}{$0.67$}& {$0.44$} \\ \cline{3-6}
RevUp                                       &                   & \multicolumn{1}{l|}{$\mathbf{0.82}$}& {$\mathbf{0.52}$} & \multicolumn{1}{l|}{$\mathbf{0.80}$}& {$\mathbf{0.63}$} \\ \hline
SSDVAE                                      & \multirow{2}{*}{0.3}                 & \multicolumn{1}{l|}{$0.60$}& {$0.28$} & \multicolumn{1}{l|}{$0.58$}& {$0.36$} \\ \cline{3-6}
RevUp                                       &                   & \multicolumn{1}{l|}{$\mathbf{0.79}$}& {$\mathbf{0.45}$} & \multicolumn{1}{l|}{$\mathbf{0.77}$}& {$\mathbf{0.58}$} \\ \hline
SSDVAE                                      &\multirow{2}{*}{0.5}                & \multicolumn{1}{l|}{$0.41$}& {$0.17$} & \multicolumn{1}{l|}{$0.41$}& {$0.23$} \\ \cline{3-6}
RevUp                                       &                  & \multicolumn{1}{l|}{$\mathbf{0.72}$}& {$\mathbf{0.36}$} & \multicolumn{1}{l|}{$\mathbf{0.71}$}& {$\mathbf{0.48}$} \\ \hline
SSDVAE                                      & \multirow{2}{*}{0.7}                 & \multicolumn{1}{l|}{$0.23$}& {$0.09$} & \multicolumn{1}{l|}{$0.22$}& {$0.11$} \\ \cline{3-6}
RevUp                                       &                  & \multicolumn{1}{l|}{$\mathbf{0.64}$}& {$\mathbf{0.29}$} & \multicolumn{1}{l|}{$\mathbf{0.63}$}& {$\mathbf{0.37}$} \\ \hline
SSDVAE                                      & \multirow{2}{*}{0.9}                  & \multicolumn{1}{l|}{$0.02$}& {$0.00$} & \multicolumn{1}{l|}{$0.02$}& {$0.00$} \\
\cline{3-6}
RevUp                                       &                  & \multicolumn{1}{l|}{$\mathbf{0.41}$}& {$\mathbf{0.09}$} & \multicolumn{1}{l|}{$\mathbf{0.25}$}& {$\mathbf{0.06}$} \\ 
\hline
\end{tabular}
}
\caption{Effect of Noise on robustness. We present standard deviations in \Cref{app:addExperiments}.}
\label{tab:effNoise}
\end{table}

\begin{table}[t]
\centering
\small
\begin{tabular}{l|l|l|l}
\hline
\multicolumn{1}{c|}{model} & \multicolumn{1}{c|}{$\epsilon$}                  & Wiki  & NYT   \\ \hline
RNNLM                      &\multicolumn{1}{c|}{-}                                          & 56.96 & 64.57 \\ \cline{3-4} 
HAQAE                      & \multicolumn{1}{c|}{-}                                     & 39.47 & 50.10 \\ \hline
SSDVAE                     & \multicolumn{1}{c|}{\multirow{2}{*}{0.0}} & 39.75 & 47.50 \\ \cline{3-4} 
RevUp                      & \multicolumn{1}{c|}{}                     & $\mathbf{39.48}$ & $\mathbf{45.36}$ \\ \hline
SSDVAE                     & \multicolumn{1}{c|}{\multirow{2}{*}{0.1}} & 39.73 & 45.91 \\ \cline{3-4} 
RevUp                      & \multicolumn{1}{c|}{}                     & $\mathbf{33.34}$ & $\mathbf{44.87}$ \\ \hline
SSDVAE                     & \multirow{2}{*}{0.7}                      & 36.79 & 44.79 \\ \cline{3-4} 
RevUp                      &                                           & $\mathbf{33.20}$ & $\mathbf{41.74}$ \\ \hline
SSDVAE                     & \multirow{2}{*}{1.0}                      & 30.69 & 36.96 \\ \cline{3-4} 
RevUp                      &                                           & $\mathbf{28.33}$ & $\mathbf{34.85}$ \\ \hline
\end{tabular}
\caption{Test perplexity results, varying the percent $\epsilon$ of side knowledge that was observed during training. We present standard deviations in \cref{app:addExperiments} (\cref{tab:full-results-ppx}).}
\label{tab:testppx}
\end{table}

\subsection{Effect of Incomplete Knowledge}
We consider the case of semi-supervised learning, where with probability $\epsilon$ a node $t_m$ is observed.
We report the classifications on the latent nodes as
$
\infdiv{p_{\mathcal{D}}{(\bt\vert\bx)}}{r_{\gamma}(\bt\vert\bx)}
\approx
-\mathbb{E}_{p_{\mathcal{D}}(\bx)p_{\theta}(\bz\vert\bx)p_{\mathcal{D}}(\bt\vert\bx)}\log r_{\gamma}(\bt\vert\bz).
$
The results are shown in \Cref{fig:Classification}. We report two widely-used classification metrics including accuracy and F1 to evaluate the performance of all methods. SSDVAE just relies on the guidance and classification but RevUp also uses the updating phase to shift the available knowledge from side information into latent variables. Thus the results demonstrate the superiorities of RevUp to predict knowledge when they are partially observed, attributable to its novel information injection and learning. 


To investigate whether the proposed approach works for task representation, we compare the perplexity of our approach to prior work in \Cref{tab:testppx}. %
Perplexity has been commonly used in the literature, which allows us to provide a fair comparison with previous efforts. %
We investigate the effect of supervision with various observation probabilities $\epsilon=$ 0.0 (unsupervised), 0.1, 0.7 and 1.0 (fully supervised) on the generated samples from the model. We see that our model is able to obtain lower perplexity scores than the previous event modeling methods. We observe as $\epsilon$ increases, the performance of our proposed model improves.  For each observation probability, our method outperforms SSDVAE. %
The results demonstrate that our method  achieves state-of-the-art performance with large margins from the baselines.

 
\section{Related Work}
\paragraph{Information Bottleneck}
The concept of using side information for discrete source coding was explored by \citet{wyner1975side,wyner1976rate}. The Information Bottleneck (IB) principle was then introduced by \citep{tishby100information} to compress input variables while predicting a target. \citet{chechik2002extracting} proposed incorporating negative side information through an auxiliary loss in a supervised manner. Our method stands apart by handling both supervised and semi-supervised settings with ease.
The Variational Information Bottleneck (VIB) was introduced by \citet{alemi2016deep}, which improved the IB estimation through amortized variational methods. \citet{voloshynovskiy2020variational} extended the VIB method for semi-supervised classification. In relation to our work, there have been numerous studies focused on maximizing the mutual information between different views and discarding the non-shared information~\citep{federici2019learning,wan2021multi,wang2019deep,mao2021deep,yan2019shared}.

 
\paragraph{Variational Autoencoders (VAEs)}
\citet{kingma2013auto} and \citet{rezende2014stochastic} introduced the reparameterization technique for variational inference. Recently, \citet{huang2021regularized}  applied adversarial training to enhance the ability of VAE in processing sequential data, while \citet{qiu2020interpretable} explored the possibility of VAE handling multi-view temporal data. Our focus is limited to objectives that can be expressed with discrete latent variables.

\Citet{jang2016categorical} and \citet{maddison2016concrete} independently presented the Gumbel-Softmax distribution, which allows backpropagation through latent discrete variables in variational autoencoders. For more information, refer to the review by \citet{huijben2022review}. \citet{van2017neural} proposed VQ-VAE, which uses vector quantization to represent discrete values with embeddings.

\citet{kingma2014semi} proposed a VAE for semi-supervised classification, where labels in unsupervised settings are treated as latent variables and predicted when available. \Citet{joy2022learning} later improved this approach by separating latent variables to differentiate label characteristics from values. \Citet{lo2020co} modified the VAE objective to incorporate external knowledge with embedding vectors, and \citet{feng2020shot} proposed a new ELBO to incorporate label prediction loss.

\paragraph{Event Modeling}
In recent years, much research has been dedicated to the challenge of modeling the sequence of events \citep{weber2018hierarchical,rezaee2021event,gao-etal-2022-improving}. One such contribution was made by \citet{weber2018event}, who introduced a tensor-based composition to effectively capture semantic event relations.
\citet{gao-etal-2022-improving} proposed a self-supervised contrastive learning approach based on co-occurring events. %
Another approach suggested by \citet{weber2018hierarchical} involves the use of Recurrent Neural Networks (RNNs) to model the hierarchical latent structures that exist in sequences of events. This methodology was further developed in a subsequent study by \citet{rezaee2021event}, where external knowledge was incorporated into the latent layer for enhanced modeling, which is highly relevant to our proposed approach. %

\section{Conclusion}
We show how to incorporate noisy partially observed side knowledge source along with latent variables. To do so, we generalized the main idea of parameter injection and maximizing the mutual information between external knowledge and latent variables. Our experiments show that our approach is more robust to noisy knowledge and outperform other baselines for the event modeling task

\section*{Acknowledgments}

We would like to thank the anonymous reviewers for their comments, questions, and suggestions. %
This material is based in part upon work supported by the National Science Foundation under Grant No. IIS-2024878. %
Some experiments were conducted on the UMBC HPCF, supported by the National Science Foundation under Grant No. CNS-1920079. %
This material is also based on research that is in part supported by the Army Research Laboratory, Grant No. W911NF2120076, and by the Air Force Research Laboratory (AFRL), DARPA, for the KAIROS program under agreement number FA8750-19-2-1003. The U.S. Government is authorized to reproduce and distribute reprints for Governmental purposes notwithstanding any copyright notation thereon. The views and conclusions contained herein are those of the authors and should not be interpreted as necessarily representing the official policies or endorsements, either express or implied, of the Air Force Research Laboratory (AFRL), DARPA, or the U.S. Government. %

\section{Limitations}
Some limitations of our work are that multiple hyper parameters should be tuned for each task. While we provide guidance and insights into effective settings (and some intuition as to why), we acknowledge that the settings may be domain dependent. 

Because we use semantic frames as our side knowledge, our focus is on improving the representation and use of discrete latent variables. While current NLP approaches have often focused on text-to-text methods for input and output, and individual words in text can be considered a form of discrete latent variables, we note that these methods are driven by large continuous embedding methods. While we believe this work can be extended to continuous cases, the approximations did make use of aspects of discrete variables, and they would need to be re-derived.

RevUp depends on mixing in statistics and learned representations from external side knowledge. While we envision this side knowledge as containing useful generalizations and semantic information, such resources could encode overly broad generalizations or other biases. While the degree of this mixture can be adjusted, imperfections or biases in the external knowledge could be captured and propagated through RevUp.

We focus on the task of event modeling, but we believe RevUp represents a step towards improving settings where noisy side information is available.

\appendix

\bibliography{refs}

\bibliographystyle{acl_natbib}

\clearpage
\appendix

\section{Information Bottleneck Principle (IB)}
Given its importance to our effort, we restate the information bottleneck principle, and discuss how it ties in with our loss formulation. Then, in subsequent appendices (\cref{app:derivation,app:updateStep,app:approxLoss}) we step through individual subloss terms.

Given the original input data $\bx$ with the target $\by$, IB aims to extract a compact representation of $\bx$ that is most informative about the target $\by$ via maximizing $\mut(y;z)-\beta \mut(x;z)$.
The first term $\mut(y;z)$ motivates the model to predict $y$, whilst the second term $\mut(x;z)$ aims at discarding irrelevant information from the input $x$. The hyperparameter $\beta$ can be set or tuned to adjust the relative importance of discarding irrelevant information. %
In our framework, we have a generalized version of IB as follows
\begin{align}
    \mathcal{L}=
    & -\mut(\by;\bt)
    +\alpha \mut(\bx;\bt\vert\bz) \nonumber \\
     & +\beta \mut(\bx;\bz)
    +\zeta \infdiv{r_{\gamma}(\bt\vert\bz)}{\mathcal{U}(\bt)}.
\end{align}
Our generalized version accounts for additional random variables, and the associated dependencies. Here, $r_\gamma(\bt\vert\bz)$ is the proposed distribution, $\mathcal{U}$ is a uniform distribution over $\bt$. The KL term helps avoid overfitting.

\section{Regularization and Task Representation Derivation}
\label{app:derivation}
Throughout we use the following inequality
\begin{align}
    &  \mathbb{E}_{p(\bz,\bx)}\log 
    \dfrac{q(\bz)}{p_{\theta}(\bz\vert\bx)}
     \leq \notag \\
    & \quad\quad\quad\quad \mathbb{E}_{p_{\mathcal{D}(\bx)}}\log
    \underbrace{
    \mathbb{E}_{p_\theta}(\bz\vert\bx) \dfrac {q(\bz)}{p_{\theta}(\bz\vert\bx)}
    }_{1}
    = 0.
    \label{app:ineq}
\end{align}
The regularization term $\mut(\bx;\bz)$ is 
\begin{align}
  \mut(\bx;\bz)  
  &=
  \mathbb{E}_{p(\bz,\bx)}\log \dfrac {p_{\theta}(\bz\vert\bx)}{p(\bz)}
  \nonumber
  \\
  &
  \leq
  \mathbb{E}_{p(\bz,\bx)}\log \dfrac {p_{\theta}(\bz\vert\bx)}{q(\bz)}
  \label{app:xzineq}
\end{align}
where since $p(\bz)=\int p_{\theta}(\bz\vert \bx)p_{\mathcal{D}}(\bx) \,dx$ is intractable we use the inequality in \Cref{app:ineq}.
Now, we can approximate \Cref{app:xzineq}
as follows
\begin{align}
  & \mathbb{E}_{p_{\mathcal{D}}(\bx)p_{\theta}(\bz\vert\bx)}  \log \dfrac {p_{\theta}(\bz\vert\bx)}{p(\bz)} \\
  &=
  \mathbb{E}_{p_{\mathcal{D}}(\bx)p_{\theta}(\bz\vert\bx)}
  \log \dfrac {\prod_{m=1}^{M}p_{\theta}(z_m\vert t_{m-1},\bx)}{\prod_{m=1}^{M}q(z_m)}
    \nonumber
  \\
  &=
  \sum_{m=1}^{M}
  \mathbb{E}_{p_{\mathcal{D}}(\bx)p_{\theta}(\bz\vert\bx)}
  \log \dfrac {p_{\theta}(z_m\vert t_{m-1},\bx)}{q(z_m)}
    \nonumber
  \\
  &
  \approx
  \dfrac{1}{S}
  \sum_{m=1}^{M}
  \sum_{s=1}^{S}
  \infdiv{p_{\theta}(z_m\vert t^{(s)}_{m-1},\bx)}{q(z_m)},
  \label{app:klapprox}
\end{align}
where
for simplicity we assume that $q(\bz)$ factorizes over independent random variables
as
$
q(\bz)
=
q(z_1,z_2,\ldots,z_M)
=\prod_{m=1}^{M}
q(z_m).
$
We parameterize each $z_m$ via a multivariate Normal, $q(z_m)=\mathcal{N}(\mathbf{0},\mathbf{I})$, and 
$p_{\theta}(z_m\vert z^{(s)}_{m-1},\bx)= \mathcal{N}(\mu_m,\Sigma_m)$:
\begin{align*}
\mu_m=f^{\mu}(z^{(s)}_{m-1},\bx) 
\\
\Sigma_m=f^{\Sigma}(z^{(s)}_{m-1},\bx),
\end{align*}
where both $f^{\mu}$ and $f^{\Sigma}$ are neural networks. With this formulation, we can calculate the KL-divergence terms of \Cref{app:klapprox} in closed forms. 

\subsection{Approximating $\mut(\by; \bt)$ (\cref{eq:taskRecons})}
\label{sec:app:I_y_t}
In the same way for the task representation, we have
\begin{align}
-\mut(\by;\bt)
  &=
  -\mathbb{E}_{p(\by,\bz,\bt)}
  \log{
  \dfrac{
  p(\by\vert\bt)
  }
  {p(\by) }
  }
  \nonumber\\
  &\leq
  -\mathbb{E}_{p(\by,\bz,\bt)}
  \log{
  \dfrac{
  q_{\phi}(\by\vert\bt)
  }
  {p(\by) }  
  }
  \nonumber\\
  &
  \leq
-\mathbb{E}_{p(\by,\bz,\bt)}
\log
 q_{\phi}(\by\vert\bt)
 +H(\by),
  \end{align}
 where $H(\by)$ is constant and we ignore it. 
 We have:
 \begin{align}
 & \mathbb{E}_{p(\by,\bz,\bt)}
\log
 q_{\phi}(\by\vert\bt) \nonumber \\
 & =\mathbb{E}_{
 p_{\mathcal{D}}{(\bx,\by)}
 p_{\theta}(\bz\vert\bx)
 p_{\gamma}(\bt\vert\bz)
 }
\log
 q_{\phi}(\by\vert\bt),
 \\
 &\approx
 \dfrac{1}{S_z}
 \dfrac{1}{S_t}
 \sum_{s_z=1}^{S_z}
  \sum_{s_t=1}^{S_t}
  \log q_{\phi}(\by\vert\bt^{(s_t)}),
 \end{align}
 where we first sample $\bx,\by \sim p_{\mathcal{D}}(\bx,\by)$, then 
 $\bz^{(s_z)} \sim p_{\theta}(\bz\vert\bx)$
 and finally $\bt^{(s_t)} \sim r_{\gamma}(\bt\vert\bz^{(s_z)})$. In this approximation $S_z$ and $S_t$ are the total number of samples for $\bz$ and $\bt$ respectively.

We wish to clarify why this term is included in the objective and how optimizing it affects the model parameters. %
The answer is that in computing $\mut(\bf y;\bf t)$, the distribution used to compute the mutual information's expectation depends on the proposed distribution $r_\gamma$. This is by construction and irrespective of whether side information is present or not. This is where the model parameters come in.

Specifically, while by definition $\mut(\bf y;\bf t) = \mathbb{E}_{p(\bf y,\bf t)}\frac{log{p(\bf y\vert\bf t)}}{p(\bf y)}$, because $p(\bf | \bf t)$ does not depend on a (free) $\bf z$, we can compute $mut(\bf y;\bf t) = \mathbb{E}_{p(\bf y,\bf z, \bf t)}  \frac{log{ p(\bf y\vert\bf t)}}{ p(\bf y) }$. %
This follows from the fact that for random variables $u$ and $v$, $\mathbb{E}_{p(u, v)}[f(u)] = \mathbb{E}_{p(u)}[f(u)]$. %
With this, we can then write
\begin{align}
  & \mut(\mathbf{y};\mathbf{t}) 
  = 
  \mathbb{E}_{p(\mathbf{y},\mathbf{t})}\log \dfrac {p(\mathbf{y}\vert\mathbf{t})}{p(\mathbf{y})}
  \nonumber
 \\
 &=
 \mathbb{E}_
{p(\mathbf y,\mathbf t,\mathbf x,\mathbf z)
}
\log\dfrac{p(\mathbf y\vert\mathbf t)}{p(\mathbf y)}
  \\
    &= 
\mathbb{E}_
{\big\{p_{\mathcal{D}}(\mathbf x,\mathbf y)
p(\bf z, \bf t | \bf x)\big\}}
\log\dfrac{p(\mathbf y\vert\mathbf t)}{p(\mathbf y)}
\\
  &= 
\mathbb{E}_
{\big\{p_{\mathcal{D}}(\mathbf x,\mathbf y)
\prod_{m=1}^{M}p_{\theta}(z_m|t_{m-1},\mathbf x){\color{red}p_{\gamma}(t_m\vert x_m,z_m;l_m)}\big\}}
\\
&\log\dfrac{p(\mathbf y\vert\mathbf t)}{p(\mathbf y)}.
\end{align}
This red term, ${\color{red}p_{\gamma}(t_m\vert x_m,z_m;l_m)}$, is defined as the interpolation of a proposal distribution $r_\gamma(t_m | z_m)$ and an empirical distribution $t_m | x_m$. Defined as a GumbelSoftmax, this proposal distribution explicitly allows uncertainty. Therefore, even when side information is present, we sample the value of ${\color{red}t_m}$ to approximate that final expectation.

Finally, as mentioned in the paper, this last line is intractable, so we learn an approximation $q_\phi(y | t)$, which introduces additional model parameters to learn. We compute $q_\phi$ using the sampled values ${\color{red}t_1...t_M}$.
     
\section{Update Step Upperbound}
\label{app:updateStep}
In this section, we show how we approximate the update term $\mut(\bx;\bz\vert\bz)$ in \Cref{eq:condMutApprox}. We have the following
\begin{align}
  \mut(\bt;\bx\vert\bz)  
  &=
  \mathbb{E}_{p_{\gamma}(\bt,\bz,\bx)}\log \dfrac {p_{\gamma}(\bt,\bx\vert\bz)}{p(\bt\vert\bz)p(\bx\vert\bz)}
  \nonumber
  \\
  &=
  \mathbb{E}_{p_{\gamma}(\bt,\bx,\bz)}\log \dfrac {p_{\gamma}(\bt\vert\bx,\bz)p(\bx\vert\bz)}{p{(\bt\vert\bz)}p(\bx\vert\bz)}
  \nonumber
  \\
  &=
  \mathbb{E}_{p_{\gamma}(\bt,\bx,\bz)}\log \dfrac {p_{\gamma}(\bt\vert\bx,\bz)}{p(\bt\vert\bz)}
  \label{app:ptvertz}
  \\
  &
  \leq
  \mathbb{E}_{p_{\gamma}(\bt,\bx,\bz)}\log \dfrac {p_{\gamma}(\bt\vert\bx,\bz)}{r_{\gamma}(\bt\vert\bz)},
    \label{app:rtvertz}
\end{align}
First, we should note that in \Cref{app:ptvertz}, the distribution $ p(\bt\vert\bz)$ is not equal to $r_{\gamma}(\bt\vert \bz) $, because for each node $m$, we have
\begin{align}
p(t_m\vert z_m)
&=
\int
p(t_m,x_m\vert z_m)\,dx_m
\nonumber
\\
&=
\int
p(t_m\vert x_m,z_m)
p(x_m\vert z_m)
\,dx_m
\nonumber
\\
&=
\int
\hat{\lambda}_mr_{\gamma}(t_m\vert z_m)p(x_m\vert z_m)\,dx_m
+ \nonumber \\
& \int
(1-\hat{\lambda}_m)p_{\mathcal{D}}(t_m\vert x_m)
p(x_m\vert z_m)
\,dx_m
\nonumber
\\
&\neq r_{\gamma}(t_m\vert z_m).
\end{align}
Our the last step of estimation in \Cref{app:rtvertz} is 
\begin{align}
  \mathbb{E}_{p_{\gamma}(\bt,\bx,\bz)}\log \dfrac {p_{\gamma}(\bt\vert\bx,\bz)}{p(\bt\vert\bz)}
  \leq \nonumber\\
  \mathbb{E}_{p_{\gamma}(\bt,\bx,\bz)}\log \dfrac {p_{\gamma}(\bt\vert\bx,\bz)}{r_{\gamma}(\bt\vert\bz)}.
  \label{app:condineq}
\end{align}
The proof for \Cref{app:condineq} is as follows.
\begin{align*}
  \mathbb{E}_{p(\bt,\bx,\bz)} &\log \dfrac {r_\gamma(\bt\vert\bz)}{p(\bt\vert\bz)} \\
  & =
  \mathbb{E}_{p(\bt,\bz)}\log \dfrac {r_{\gamma}(\bt\vert\bz)}{p(\bt\vert\bz)}
  \nonumber
  \\
  &=
  \mathbb{E}_{p(\bz)}{\mathbb{E}_{p(\bt\vert\bz)}\log \dfrac {r_{\gamma}(\bt\vert\bz)}{p(\bt\vert\bz)}}
  \\
  &\leq
  \mathbb{E}_{p(\bz)}\log
  \big\{
  \underbrace{\mathbb{E}_{p(\bt\vert\bz)} \dfrac {r_{\gamma}(\bt\vert\bz)}{p(\bt\vert\bz)}}_{1}
  \big\}
  \\
  &
  =
  0
\end{align*}

\section{Update State Approximation for Discrete Case}
  \label{app:approxLoss}
We show how we approximate the upperbound presented in \Cref{app:rtvertz}.
\begin{align}
  & \mathbb{E}_{p_{\gamma}(\bt,\bz,\bx)}\log \dfrac{p_{\gamma}(\bt\vert\bz,\bx)}{r_{\gamma}(\bt\vert\bz)}
      \nonumber \\
  & =
  \underbrace{
  \mathbb{E}_{p_{\mathcal{D}}(\bx)p_{\theta}(\bz\vert\bx)p_{\gamma}(\bt\vert\bz,\bx)}\log {p_{\gamma}(\bt\vert\bz,\bx)}
  }_{\circled{$\text{A}_1$}} \nonumber \\
  & -
  \underbrace {
  \mathbb{E}_{p_{\mathcal{D}}(\bx)p_{\theta}(\bz\vert\bx)p_{\gamma}(\bt\vert\bz,\bx)}\log {r_{\gamma}(\bt\vert\bz)} 
  }_
  {\circled{$\text{A}_2$}}
  .
\end{align}
Here ${\circled{$\text{A}_1$}}$ is the negative entropy of the revised distribution $p_{\gamma}(\bt\vert\bx,\bz)$:
\begin{align}
    {\circled{$\text{A}_1$}} 
    &=
    -H(\bt\vert\bx,\bz)
    \\
    &=- \sum_{m=1} H(t_m\vert x_m,z_m).
\end{align}
For the second term we have:
\begin{align}
  {\circled{$\text{A}_2$}}
  &=
  \mathbb{E}_{p_{\mathcal{D}}(\bx)}
  \mathbb{E}_{p_{\theta}(\bz\vert\bx)}
  \left[
  \hat{\lambda}
  \mathbb{E}_{r_{\gamma}(\bt\vert\bz)}
  + \right.\nonumber \\
  & \quad\quad\quad\left.(1-\hat{\lambda})
  \mathbb{E}_{p_{\mathcal{D}}(\bt\vert\bx)}
  \right]
  \log {r_{\gamma}(\bt\vert\bz)} 
  \nonumber
  \\
  &=
  \hat{\lambda}
  \underbrace{
  \mathbb{E}_{p_{\mathcal{D}}(\bx)}
  \mathbb{E}_{p_{\theta}(\bz\vert\bx)}
  \mathbb{E}_{r_{\gamma}(\bt\vert\bz)}
  \log {r_{\gamma}(\bt\vert\bz)} 
  }_{\circled{$\text{A}_{21}$}} \nonumber \\
  & +
  (1-\hat{\lambda})
  \underbrace{
  \mathbb{E}_{p_{\mathcal{D}}(\bx)}
  \mathbb{E}_{p_{\mathcal{D}}(\bt\vert\bx)}
  \log {r_{\gamma}(\bt\vert\bz)} 
  }_{\circled{$\text{A}_{22}$}}.
\end{align}

Similarly ${\circled{$\text{A}_{21}$}}$ is the entropy of the proposed distribution $r_{\gamma}(\bt \vert \bz)$:
\begin{align}
    {\circled{$\text{A}_{21}$}}&=-H(\bt\vert\bz)
    \\
    &=-\sum_{m=1}^{M} H(t_m\vert z_m)
\end{align}
Also ${\circled{$\text{B}_{22}$}}$ guides the model to discriminator $\bt$ given $\bz$ which is a classifier.
To compute the entropy terms, in the discrete case (Gumbel Softmax), 
we use Monte Carlo sampling as follows
\begin{align}
H(t_m\vert x_m,z_m)
& \approx
  \dfrac{1}{S}
  \sum_{n=1}^{N}
  \sum_{s=1}^{S}
  \hat{\boldsymbol{\gamma}}^{\top(s)}_m
  \log
  \hat{\boldsymbol{\gamma}}^{(s)}_m, \\
H(t_m\vert z_m)
&    \approx
  \dfrac{1}{S}
  \sum_{s=1}^{S}
  \boldsymbol{\gamma}^{\top(s)}_m\log \boldsymbol{\gamma}^{(s)}_m,
\end{align}
where $S$ is the number of point estimates and 
$\hat{\gamma}_{m}$ is derived from \Cref{eq:lambdaHat}. %
Here, we have defined $\hat{\boldsymbol{\gamma}}$ to be 
$[
\hat{\lambda}
\gamma_1,
\hat{\lambda}
\gamma_2,
\ldots,
\hat{\lambda}
\gamma_{k^*}+(1-\hat{\lambda}),
\ldots,
\hat{\lambda}
\gamma_K
].$
The $\hat{\boldsymbol{\gamma}}$ can be thought of as a mixture of the $\boldsymbol{\gamma}$ and empirical categorical distributions. %
(We use this definition of $\hat{\boldsymbol{\gamma}}$ in the proof of \cref{thmGeneralize} in \cref{app:ssdvaeGenProof}.) %
To sample from the revised distribution $p_{\gamma}(t_m|\bx,\bz;l_m)$ we first draw a Bernoulli sample 
  $
 \hat\lambda_m 
 \sim \text{Bern}(\hat\lambda_m; 1/(1+\lambda l_m))
 $ and consequently
we draw knowledge sample from the mixture of probabilities by
$
t_m \sim
  \hat{\lambda}_m r_{\gamma}(t_m\vert z_m) +(1-\hat{\lambda}_m) p_{\mathcal{D}}(t_m\vert x_m).
  $
  We implement this with a single-sample approximation.

\section{Revision Step Property of $p_{\gamma}
  \big(
    t_m\vert x_m,z_m;l_m
  \big)$}
\label{app:dropOut}
\begin{figure}
\centering
\includegraphics[width=0.35\textwidth]{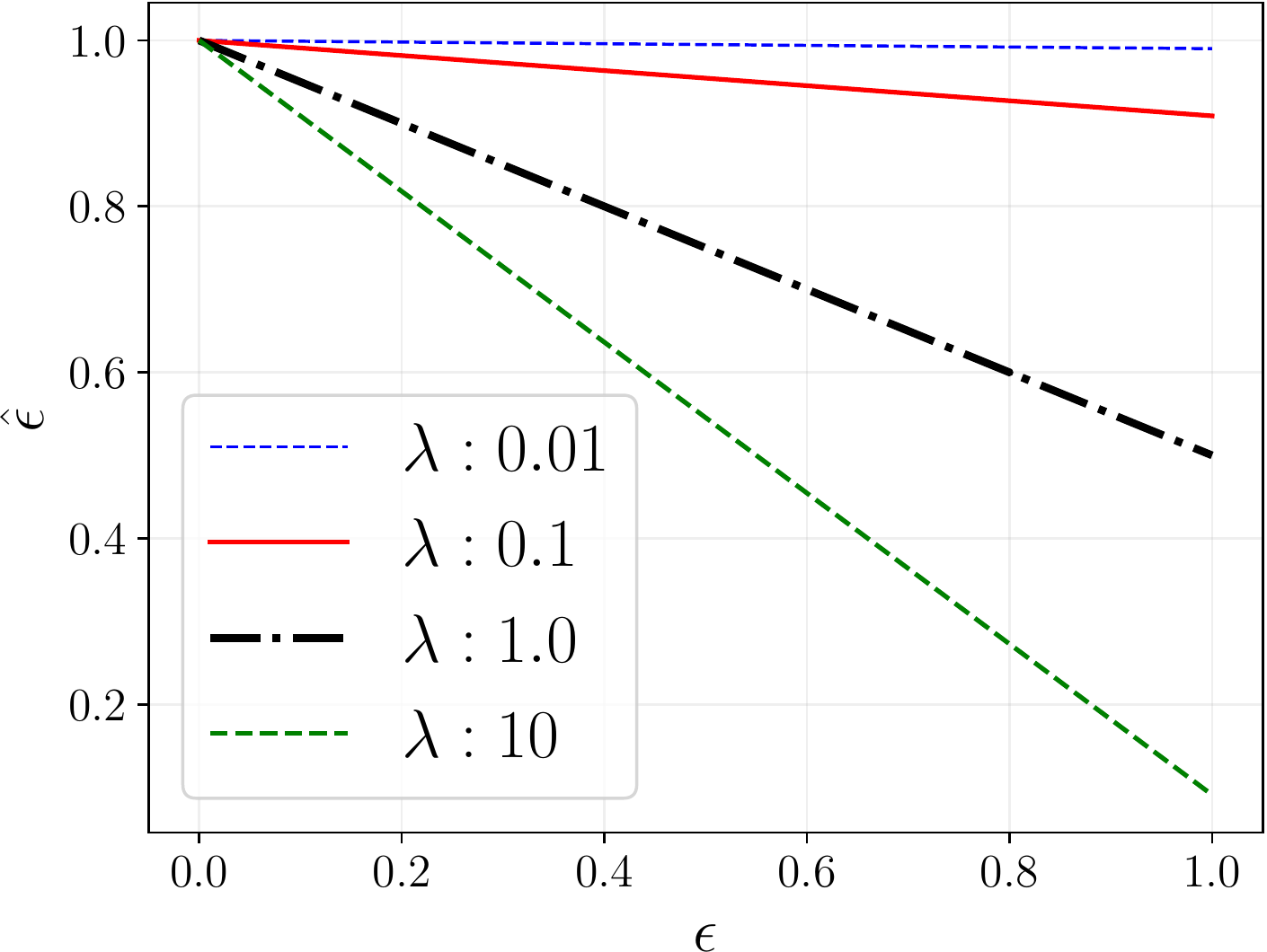}
\caption{Effect of $\lambda$ and observation probability $\epsilon$ on $\hat{\epsilon}$}
\label{fig:hatEps}
\vspace{2ex}
\end{figure}

In this section we show how the formulation of $p_{\gamma}\left(t_m\vert x_m,z_m;l_m\right)$ 
leads to the property that, during the revision step, we rely more on the available side/external knowledge as more of it becomes available. %
Recall that we initially defined
$p_{\gamma}\left(t_m\vert x_m,z_m;l_m\right)
=
  \dfrac
  {1}{1+\lambda l_m}
  r_{\gamma}(\st_m\vert z_m)
  +
\dfrac{
  \lambda
  l_{m}
  }
  {
  1
  +
  \lambda
  l_{m}
  }
p_{\mathcal{D}}{(\st_m\vert x_m)},$
where $\lambda$ allows a fractional portion of the empirical distribution $p_{\mathcal{D}}$ to influence the learned model. %
Suppose that $l_{m}\sim \mathrm{Bern}(\epsilon)$ and 
$\tilde{\lambda}=
1/(1+\lambda)
$, given node $m$ we have
\begin{align}
\begin{split}
  & p
  \big(
    t_m\vert x_m,z_m
  \big)
  =
  \sum_{l_m}
  p
  \big(
    t_m\vert x_m,z_m,l_m
  \big)
  p
  (
  {l}_m
  )
  \nonumber\\
  &=
  \epsilon p_{\gamma}{(t_m\vert x_m,z_m,l_m)}+(1-\epsilon)r_{\gamma}(t_m\vert z_m)
  \nonumber\\
  &=
  \epsilon
  \left[
  \tilde{\lambda}
  r_{\gamma}(t_m\vert z_m)
  + (1-\tilde{\lambda})
  p_{\mathcal{D}}{(t_m\vert x_m)}
  \right] \\
  & \quad\quad\quad
  + 
  \big(1-\epsilon\big)
  r_{\gamma}(t_m\vert z_m)
  \nonumber\\
 &=
  \big(1-\epsilon(1-\tilde{\lambda})\big)
  r_{\gamma}(t_m\vert z_m) \\
    & \quad\quad +  \epsilon(1-\tilde{\lambda})
  p_{\mathcal{D}}{(t_m\vert x_m)}
  \nonumber \\
  &=
  \hat{\epsilon}
  r_{\gamma}(t_m\vert z_m)
  +
  (1-\hat{\epsilon})
  p_{\mathcal{D}}{(t_m\vert x_m)},
\end{split}
\end{align}
where 
$
\hat{\epsilon}=1-\epsilon
\big(\dfrac{\lambda}{1+\lambda}\big).
$
In particular, this formulation allows us to account for both our estimate of observation of side knowledge, and the extent to which we wish to use it when learning our model. %
In \Cref{fig:hatEps}, we show the effect of $\lambda$ and $\epsilon$ on $\hat{\epsilon}.$ %



\section{Generalization of SSDVAE (Proof of \Cref{thmGeneralize})}
\label{app:ssdvaeGenProof}
We show that under specific setting, our model reduces to the SSDVAE parameter injection mechanism.
\begin{thmwonumb}
If $r_{\gamma}(\st\vert z)=\text{Cat}(\boldsymbol\gamma)$  , a categorical distribution with parameters $\gamma$, and the empirical distribution
$p_{\mathcal{D}}{(t\vert x)}$ is a one-hot representation with $t_{k^*}=1$, the revision step reduces to SSDVAE parameter injection.
\end{thmwonumb}
\begin{proof}
\normalfont {
Using the interpolation form
$p_{\gamma}
  \big(
    t_m\vert x_m,z_m;l_m
  \big)
  = 
  \hat{\lambda}_m
    r_{\gamma}(t_m\vert z_m)
    + 
  (1-\hat{\lambda}_m)
  p_{\mathcal{D}}{(\st_m\vert x_m)}$ %
and due to the fact that the mixture of categorical distributions is a single categorical, 
we have
}
$
  p
  \big(
    t\vert x, z;l=1
  \big)
  =
  \mathrm{Cat}(\hat{\boldsymbol \gamma}),
$
where
\begin{equation}
\hat{\boldsymbol \gamma}=
[
\hat{\lambda}
\gamma_1,
\hat{\lambda}
\gamma_2,
\ldots,
\hat{\lambda}
\gamma_{k^*}+(1-\hat{\lambda}),
\ldots,
\hat{\lambda}
\gamma_K
].
\label{eq:lambdaHat}
\end{equation}
Given $0\leq\gamma_i\leq 1$ and $0\leq\hat{\lambda}\leq 1$,
we have $\hat{\lambda}\gamma_i \leq  \gamma_i ~(\forall i\neq k^*)$, and $\hat{\lambda}\gamma_{k^*} +(1-\hat{\lambda}) \geq \gamma_{k^*} $.
This implies that for all the indices $i\neq k^*$, the value of  $\gamma_i$ is reduced and $\gamma_{k^*}$ is increased,
confirming that parameter injection is a specific form of our definition.
\end{proof}

\section{Updating Step Maximizes Mutual Information (Proof of \Cref{theoremMut})}
  \label{app:mutalInfo}
  We show that for the reconstruction tasks, where $\by$ is a copy of $\bx$, minimizing $\mut(\bx;\bt\vert\bz)$ maximizes $\mut(\bt;\bz)$.
  While not explicitly reﬂected in our model definition, note that $\mut(\bt;\bz)$ is intractable. This can be seen in the definition of $\mut(\bt;\bz)$,
\begin{align}
\mut(\bt;\bz) = \mathbb{E}_{p(\bt,\bz)}
\log \dfrac{p(\bt\vert\bz)}{p(\bt)},
\end{align}
here both $p(\bt,\bz)$ and $p(\bt)$ are intractable due to the integral over $\bx$. 
Following \citet{cover1999elements}, the mutual information between three random variables $\bx$, $\bt$, and $\bz$ can be deﬁned as  
  \begin{align}
      \mut(\bx;\bt;\bz)&=\mut(\bx;\bt)-\mut(\bx;\bt\vert\bz).
      \label{eq:threeMut}
  \end{align}
  \Cref{eq:threeMut} is symmetric in $\bx$, $\bt$ and $\bz$, so we have:
 \begin{align}
    \mut(\bx;\bt;\bz)&=\mut(\bt;\bz)-\mut(\bt;\bz\vert\bx).
         \label{eq:alterThreeMut}
 \end{align}
As already proven by \citet{federici2019learning}, the second term $\mut(\bt;\bz\vert \bx)$ equals to zero. Combining \Cref{eq:threeMut} and \Cref{eq:alterThreeMut}, we have
\begin{align}
   \mut(\bx;\bt)-\mut(\bx;\bt\vert\bz)=\mut(\bt;\bz)
   \label{eq:threeMutEq}
\end{align}
As evident from \Cref{eq:threeMutEq}, maximizing $\mut(\bx;\bt)$ and minimizing $\mut(\bx;\bt\vert\bz)$ is equivalent to maximizing $\mut(\bt;\bz)$.

  

\section{Full Results}
\label{app:addExperiments}
In this section, we present the full results for those presented in the main paper. Specifically, the results presented here augment those in \cref{fig:Classification}, \Cref{tab:effNoise}, and \Cref{tab:testppx}.

\paragraph{Results on the Effect of Noisy Knowledge ($\eta$)}
In \Cref{tab:full-results-noise}, we how the full results from our noisy knowledge experiments, shown in the main paper's \Cref{tab:effNoise}. %
As with our other results, these are averaged over three runs. %
In our development we noticed that precision tended to be the main hindrance on F1, and therefore we additionally report precision.

\begin{table*}
 \resizebox{1.0\textwidth}{!}{
\begin{tabular}{l|l|llllll}
\specialrule{.1em}{.05em}{.05em}
\multirow{3}{*}{model} & \multirow{3}{*}{noise $\eta$} & \multicolumn{6}{c}{Dataset}                                                                                                               \\ \cline{3-8} 
                       &                        & \multicolumn{3}{c|}{Wiki}                                                      & \multicolumn{3}{c}{NYT}                                  \\ \cline{3-8} 
                       &                        & \multicolumn{1}{c|}{Acc} & \multicolumn{1}{c|}{F1} & \multicolumn{1}{c|}{Prec} & \multicolumn{1}{c|}{Acc} & \multicolumn{1}{c|}{F1} & \multicolumn{1}{c}{Prec} \\ \hline
SSDVAE                 & \multicolumn{1}{c|}{\multirow{2}{*}{0.1}}                    & \multicolumn{1}{l|}{${0.77\pm 0.008}$}    & \multicolumn{1}{l|}{${0.43\pm 0.019}$}   & \multicolumn{1}{l|}{${0.41\pm 0.015}$}     & \multicolumn{1}{l|}{${0.76\pm 0.000}$}    &
\multicolumn{1}{l|}{${0.55\pm 0.001}$}   &  
\multicolumn{1}{l}{${0.52\pm 0.001}$}    \\ \cline{3-8} 
RevUp                  &                    & \multicolumn{1}{l|}{${\mathbf{0.85\pm 0.005}}$}    & \multicolumn{1}{l|}{$\mathbf{0.58\pm 0.015}$}   & \multicolumn{1}{l|}{$\mathbf{0.56\pm 0.012}$}     & \multicolumn{1}{l|}{$\mathbf{0.83\pm 0.000}$ }    &
\multicolumn{1}{l|}{$\mathbf{0.71\pm 0.002}$}   &    
\multicolumn{1}{l}{$\mathbf{0.69\pm 0.002}$}
\\ \hline
SSDVAE                 & \multicolumn{1}{c|}{{\multirow{2}{*}{0.2}} }                    & \multicolumn{1}{l|}{${0.69\pm 0.002}$}    & \multicolumn{1}{l|}{${0.35\pm 0.001}$}   & \multicolumn{1}{l|}{${0.33\pm 0.002}$}     & \multicolumn{1}{l|}{${0.67\pm 0.001}$  }    &
\multicolumn{1}{l|}{${0.44\pm 0.000}$ }   &
\multicolumn{1}{l}{${0.41\pm 0.001}$}
\\ \cline{3-8} 
RevUp                  &                   & \multicolumn{1}{l|}{$\mathbf{0.82\pm 0.001}$}    & \multicolumn{1}{l|}{$\mathbf{0.52\pm 0.001}$}   & \multicolumn{1}{l|}{$\mathbf{0.49\pm 0.003}$}     & \multicolumn{1}{l|}{$\mathbf{0.80\pm 0.000}$ }    &
\multicolumn{1}{l|}{$\mathbf{0.63\pm 0.001}$ }   &
\multicolumn{1}{l}{$\mathbf{0.60\pm 0.001}$}
\\ \hline
SSDVAE                 & \multicolumn{1}{c|}{{\multirow{2}{*}{0.3}} }                   & \multicolumn{1}{l|}{${0.60\pm 0.010}$}    & \multicolumn{1}{l|}{${0.28\pm 0.018}$}   & \multicolumn{1}{l|}{${0.27\pm 0.013}$}     & \multicolumn{1}{l|}{ ${0.58\pm 0.001}$}    &
\multicolumn{1}{l|}{${0.36\pm 0.000}$  }   &
\multicolumn{1}{l}{ ${0.34\pm 0.000}$ }
\\ \cline{3-8} 
RevUp                  &                & \multicolumn{1}{l|}{$\mathbf{0.79\pm 0.008}$}    & \multicolumn{1}{l|}{$\mathbf{0.45\pm 0.023}$}   & \multicolumn{1}{l|}{$\mathbf{0.42\pm 0.019}$}     & \multicolumn{1}{l|}{$\mathbf{0.77\pm 0.000}$}    &
\multicolumn{1}{l|}{$\mathbf{0.58\pm 0.001}$ }   &
\multicolumn{1}{l}{$\mathbf{0.55\pm 0.000}$}
\\ \hline
SSDVAE                 & \multicolumn{1}{c|}{{\multirow{2}{*}{0.5}} }                    & \multicolumn{1}{l|}{${0.41\pm 0.017}$}    & \multicolumn{1}{l|}{${0.17\pm 0.014}$}   & \multicolumn{1}{l|}{${0.18\pm 0.011}$}     & \multicolumn{1}{l|}{${0.41\pm 0.003}$}    &
\multicolumn{1}{l|}{${0.23\pm 0.001}$}   & 
\multicolumn{1}{l}{${0.23\pm 0.002}$}
\\ \cline{3-8} 
RevUp                  &                 & \multicolumn{1}{l|}{$\mathbf{0.72\pm 0.007}$}    & \multicolumn{1}{l|}{$\mathbf{0.36\pm 0.022}$}   & \multicolumn{1}{l|}{$\mathbf{0.34\pm 0.017}$}     & \multicolumn{1}{l|}{$\mathbf{0.71\pm 0.001}$ }    &
\multicolumn{1}{l|}{$\mathbf{0.48\pm 0.001}$}   &
\multicolumn{1}{l}{$\mathbf{0.45\pm 0.000}$ }
\\ \hline
SSDVAE                 & \multicolumn{1}{c|}{{\multirow{2}{*}{0.7}} }                    & \multicolumn{1}{l|}{${0.23\pm 0.002}$}    & \multicolumn{1}{l|}{${0.09\pm 0.001}$}   & \multicolumn{1}{l|}{${0.11\pm 0.001}$}     & \multicolumn{1}{l|}{ ${0.22\pm 0.004}$ }    &
\multicolumn{1}{l|}{${0.11\pm 0.002}$}   &
\multicolumn{1}{l}{${0.13\pm 0.002}$}
\\ \cline{3-8} 
RevUp                  &                    & \multicolumn{1}{l|}{$\mathbf{0.64\pm 0.005}$}    & \multicolumn{1}{l|}{$\mathbf{0.29\pm 0.003}$}   & \multicolumn{1}{l|}{$\mathbf{0.28\pm 0.002}$}     & \multicolumn{1}{l|}{$\mathbf{0.63\pm 0.003}$ }    &
\multicolumn{1}{l|}{$\mathbf{0.37\pm 0.002}$}   &
\multicolumn{1}{l}{$\mathbf{0.35\pm 0.002}$ }
\\ \hline
SSDVAE                 & \multicolumn{1}{c|}{{\multirow{2}{*}{0.9}} }                   & \multicolumn{1}{l|}{${0.02\pm 0.001}$}    & \multicolumn{1}{l|}{${0.00\pm 0.000}$ }   & \multicolumn{1}{l|}{${0.01\pm 0.000}$}     & \multicolumn{1}{l|}{${0.02\pm 0.001}$  }    &
\multicolumn{1}{l|}{${0.00\pm 0.000}$}   &
\multicolumn{1}{l}{ ${0.01\pm 0.000}$}
\\ \cline{3-8} 
RevUp                  &                   & \multicolumn{1}{l|}{$\mathbf{0.41\pm 0.009}$}    & \multicolumn{1}{l|}{$\mathbf{0.09\pm 0.003}$}   & \multicolumn{1}{l|}{$\mathbf{0.11\pm 0.003}$}     & \multicolumn{1}{l|}{$\mathbf{0.25\pm 0.039}$}    &
\multicolumn{1}{l|}{$\mathbf{0.06\pm 0.013}$}   &
\multicolumn{1}{l}{$\mathbf{0.09\pm 0.015}$ }
\\ 
\specialrule{.1em}{.05em}{.05em}
\end{tabular}
}
\vspace{2ex}
\caption{Effect of Noise on the test dataset.}
\label{tab:full-results-noise}
\end{table*}

\paragraph{Full Perplexity Results}
In \Cref{tab:full-results-ppx}, we how the full language modeling perplexity results that are shown in the main paper's \Cref{tab:testppx}. %
Here, we are showing results across both the dev and test sets. %
We also include standard deviation results (from three runs). %

\begin{table}[t]
\centering{
\resizebox{.98\columnwidth}{!}{
\begin{tabular}{l|c|llll|}
\specialrule{.1em}{.05em}{.05em}
\multicolumn{1}{c|}{\multirow{3}{*}{model}} & \multirow{3}{*}{$\epsilon$} & \multicolumn{4}{c}{Dataset}                                                                                    \\ \cline{3-6} 
\multicolumn{1}{c|}{}                       &                      & \multicolumn{2}{c|}{Wiki}                              & \multicolumn{2}{c}{NYT}                               \\ \cline{3-6} 
\multicolumn{1}{c|}{}                       &                      & \multicolumn{1}{c|}{Valid} & \multicolumn{1}{c|}{Test} & \multicolumn{1}{c|}{Valid} & \multicolumn{1}{c}{Test} \\ \hline
RNNLM                                        & -                        & \multicolumn{1}{l|}{64.02 $\pm 2.53$}      &   \multicolumn{1}{l|}{64.57 $\pm 2.60$}     & \multicolumn{1}{l|}{65.07 $\pm 3.25$}      &    \multicolumn{1}{l}{56.96 $\pm 2.82$}   \\ \hline
HAQAE                                       &-                        & \multicolumn{1}{l|}{49.03 $\pm 3.90$}      &    \multicolumn{1}{l|}{50.10 $\pm 4.05$}     & \multicolumn{1}{l|}{43.13 $\pm 5.29$}      &    \multicolumn{1}{l}{39.47 $\pm 4.84$}     \\ \hline
SSDVAE                                      & \multirow{2}{*}{0.0}                    & \multicolumn{1}{l|}{46.61 $\pm 1.08$}      &  \multicolumn{1}{l|}{47.50 $\pm 1.06$}      & \multicolumn{1}{l|}{44.80 $\pm 0.85$}      &  \multicolumn{1}{l}{39.75 $\pm 1.21$}       \\ \cline{3-6} 
RevUp                                       &                    & \multicolumn{1}{l|}{\textbf{44.38} $\pm \mathbf{1.38}$}      &  \multicolumn{1}{l|}{\textbf{45.36} $\pm \mathbf{1.40}$}  & \multicolumn{1}{l|}{\textbf{42.61 $\pm 0.58$}}      &  \multicolumn{1}{l}{\textbf{39.48 $\pm 0.49$}}   \\ \hline
SSDVAE                                      &\multirow{2}{*}{0.1}                      & \multicolumn{1}{l|}{45.17 $\pm 1.17$}      &  \multicolumn{1}{l|}{45.91 $\pm 1.15$}     & \multicolumn{1}{l|}{43.52 $\pm 0.16$}      &  \multicolumn{1}{l}{39.73 $\pm 0.16$}   \\ \cline{3-6} 
RevUp                                       &                      & \multicolumn{1}{l|}{\textbf{43.99} $\pm \mathbf{1.79}$}      &     \multicolumn{1}{l|}{\textbf{44.87} $\pm \mathbf{1.83}$}    & \multicolumn{1}{l|}{\textbf{36.10 $\pm 0.94$}}      &    \multicolumn{1}{l}{\textbf{33.34 $\pm 0.96$}}     \\ \hline
SSDVAE                                      & \ \multirow{2}{*}{0.7}                      & \multicolumn{1}{l|}{44.38 $\pm 0.55$}      &  \multicolumn{1}{l|}{44.79 $\pm 0.58$}     &  \multicolumn{1}{l|}{39.77 $\pm 1.05$}      &  \multicolumn{1}{l}{36.79 $\pm 0.91$}         \\ \cline{3-6} 
RevUp                                       &                  & \multicolumn{1}{l|}{\textbf{40.90} $\mathbf{\pm 1.11}$}      &  \multicolumn{1}{l|}{\textbf{41.74} $\pm \mathbf{1.05}$}    & \multicolumn{1}{l|}{\textbf{35.77 $\pm 0.25$}}      & \multicolumn{1}{l}{\textbf{33.20 $\pm 0.17$}}   \\ \hline
SSDVAE                                      &\multirow{2}{*}{1.0}                    & \multicolumn{1}{l|}{36.79 $\pm 0.33$}      & \multicolumn{1}{l|}{36.96 $\pm 0.34$}    & \multicolumn{1}{l|}{33.18 $\pm 0.39$}      &  \multicolumn{1}{l}{30.69 $\pm 0.31$} \\ \cline{3-6} 
RevUp                                       &                       & \multicolumn{1}{l|}{\textbf{34.28 $\pm \mathbf{0.89}$}}      &  \multicolumn{1}{l|}{\textbf{34.85} $\pm \mathbf{0.90}$}     & \multicolumn{1}{l|}{\textbf{30.61 $\pm \mathbf{0.38}$}}      &  \multicolumn{1}{l}{\textbf{28.33 $\pm 0.43$} }  \\ 
\specialrule{.1em}{.05em}{.05em}
\end{tabular}
}
\vspace{2ex}
\caption{Perplexity results on the Wikipedia and NYT datasets (extension of \Cref{tab:testppx}). Main results are the average of three runs, along with standard deviation.}
\label{tab:full-results-ppx}
}
\end{table}

\paragraph{Full Results on Frame Classification}
In \Cref{tab:full-classification-results}, we show the full results for the side knowledge (per-event frame) prediction experiments we reported in \Cref{fig:Classification}. The main classification results were averaged over three runs, along with standard deviation. These results are visible in \Cref{fig:Classification}, but we include the numeric values for any future comparisons.

\begin{table*}
\resizebox{.98\textwidth}{!}{
\begin{tabular}{l|c|llllll}
\specialrule{.1em}{.05em}{.05em}
\multicolumn{1}{c|}{\multirow{3}{*}{model}} & \multirow{3}{*}{$\epsilon$} & \multicolumn{6}{c}{Dataset}                                                                                                                                    \\ \cline{3-8} 
\multicolumn{1}{c|}{}                       &                        & \multicolumn{3}{c|}{Wiki}                                                      & \multicolumn{3}{c}{NYT}                                                       \\ \cline{3-8} 
\multicolumn{1}{c|}{}                       &                        & \multicolumn{1}{c|}{Acc} & \multicolumn{1}{c|}{F1} & \multicolumn{1}{c|}{Prec} & \multicolumn{1}{c|}{Acc} & \multicolumn{1}{c|}{F1} & \multicolumn{1}{c}{Prec} \\ \hline
SSDVAE                                      & \multicolumn{1}{c|}{\multirow{2}{*}{0.1}}                     &
\multicolumn{1}{l|}{${0.02\pm 0.002}$ }    & \multicolumn{1}{l|}{${0.01\pm 0.001}$  }   & \multicolumn{1}{l|}{ ${0.02\pm 0.002}$}
& \multicolumn{1}{l|}{${ 0.02\pm 0.002}$ }    & \multicolumn{1}{l|}{${ 0.00\pm 0.001}$ }   &        \multicolumn{1}{l}{${ 0.01\pm 0.003}$}  
\\ \cline{3-8} 
RevUp                                       &                    &
\multicolumn{1}{l|}{$\mathbf{0.06\pm 0.001}$ }    & \multicolumn{1}{l|}{$\mathbf{0.02\pm 0.000}$  }   &
\multicolumn{1}{l|}{ $\mathbf{0.04\pm 0.000}$}     & \multicolumn{1}{l|}{$\mathbf{ 0.05\pm 0.003}$}    & \multicolumn{1}{l|}{$\mathbf{ 0.02\pm 0.001}$ }   & 
\multicolumn{1}{l}{$\mathbf{ 0.04\pm 0.002}$} 
\\ \hline
SSDVAE                                      & \multicolumn{1}{c|}{\multirow{2}{*}{0.3}}                    & \multicolumn{1}{l|}{${0.10\pm 0.002}$}      &  \multicolumn{1}{l|}{${0.05\pm 0.001}$}   &  \multicolumn{1}{l|}{${0.08\pm 0.002}$}  & \multicolumn{1}{l|}{${ 0.11\pm 0.003}$}            & \multicolumn{1}{l|}{${ 0.06\pm 0.001}$}           & \multicolumn{1}{l}{${ 0.09\pm 0.002}$}  
\\ \cline{3-8} 
RevUp                                       &                    & \multicolumn{1}{l|}{$\mathbf{0.23\pm 0.002}$}     &  \multicolumn{1}{l|}{$\mathbf{0.11\pm 0.001}$}    &  \multicolumn{1}{l|}{$\mathbf{0.13\pm 0.000}$}&
\multicolumn{1}{l|}{$\mathbf{ 0.17\pm 0.005}$}            & \multicolumn{1}{l|}{$\mathbf{ 0.07\pm 0.002}$}           & \multicolumn{1}{l}{$\mathbf{ 0.09\pm 0.002}$} 
\\ \hline
SSDVAE                                      & \multicolumn{1}{c|}{\multirow{2}{*}{0.5}}                     & \multicolumn{1}{l|}{${0.23\pm 0.002}$}     &  \multicolumn{1}{l|}{${0.11\pm 0.000}$}    &  \multicolumn{1}{l|}{${0.14\pm 0.001}$}  & \multicolumn{1}{l|}{${ 0.25\pm 0.005}$}            & \multicolumn{1}{l|}{${ 0.15\pm 0.002}$}           & \multicolumn{1}{l}{${ 0.18\pm 0.003}$}
\\ \cline{3-8} 
RevUp                                       &                     &  \multicolumn{1}{l|}{$\mathbf{0.44\pm 0.014}$}     &  \multicolumn{1}{l|}{$\mathbf{0.22\pm 0.009}$}    &  \multicolumn{1}{l|}{$\mathbf{0.21\pm 0.006}$} &
\multicolumn{1}{l|}{$\mathbf{ 0.36\pm 0.017}$}            & \multicolumn{1}{l|}{$\mathbf{ 0.22\pm 0.008}$}           & \multicolumn{1}{l}{$\mathbf{ 0.22\pm 0.007}$}
\\ \hline
SSDVAE                                      &\multicolumn{1}{c|}{\multirow{2}{*}{0.7}}                    & \multicolumn{1}{l|}{${0.52\pm 0.008}$}     &  \multicolumn{1}{l|}{${0.28\pm 0.008}$}    &  \multicolumn{1}{l|}{${0.28\pm 0.007}$}
&
 \multicolumn{1}{l|}{${ 0.56\pm 0.005}$}            & \multicolumn{1}{l|}{${ 0.40\pm 0.005}$}            & \multicolumn{1}{l}{${ 0.39\pm 0.004}$}
\\ \cline{3-8} 
RevUp                                       &                     & \multicolumn{1}{l|}{$\mathbf{0.84\pm 0.004}$}     &  \multicolumn{1}{l|}{$\mathbf{0.65\pm 0.003}$}    &  \multicolumn{1}{l|}{$\mathbf{0.65\pm 0.002}$} &
\multicolumn{1}{l|}{$\mathbf{ 0.76\pm 0.010}$}             & \multicolumn{1}{l|}{$\mathbf{ 0.61\pm 0.015}$}           & \multicolumn{1}{l}{$\mathbf{ 0.57\pm 0.018}$} 
\\ \hline
SSDVAE                                      & \multicolumn{1}{c|}{\multirow{2}{*}{0.9}}                     & \multicolumn{1}{l|}{${0.84\pm 0.004}$}     &  \multicolumn{1}{l|}{${0.61\pm 0.008}$}    &  \multicolumn{1}{l|}{${0.62\pm 0.010}$} & \multicolumn{1}{l|}{${ 0.83\pm 0.001}$}            & \multicolumn{1}{l|}{${ 0.77\pm 0.013}$}           & \multicolumn{1}{l}{${ 0.78\pm 0.014}$}
\\ \cline{3-8} 
RevUp                                       &                    &  \multicolumn{1}{l|}{$\mathbf{0.86\pm 0.001}$}     &  \multicolumn{1}{l|}{$\mathbf{0.72\pm 0.003}$}    &  \multicolumn{1}{l|}{$\mathbf{0.73\pm 0.005}$}&
\multicolumn{1}{l|}{$\mathbf{ 0.85\pm 0.002}$}            & \multicolumn{1}{l|}{$\mathbf{ 0.81\pm 0.007}$}           & \multicolumn{1}{l}{$\mathbf{ 0.82\pm 0.008}$}  
\\ \hline
SSDVAE                                      &\multicolumn{1}{c|}{\multirow{2}{*}{1.0}}                     & \multicolumn{1}{l|}{${0.85\pm 0.002}$}     &  \multicolumn{1}{l|}{${0.68\pm 0.003}$}    &  \multicolumn{1}{l|}{${0.69\pm 0.003}$}
&
\multicolumn{1}{l|}{${ 0.84\pm 0.001}$}            & \multicolumn{1}{l|}{${ 0.81\pm 0.002}$}           & \multicolumn{1}{l}{${ 0.82\pm 0.001}$}
\\ \cline{3-8} 
RevUp                                       &                    &  \multicolumn{1}{l|}{$\mathbf{0.87\pm 0.003}$}     &  \multicolumn{1}{l|}{$\mathbf{0.75\pm 0.008}$}   &  \multicolumn{1}{l|}{$\mathbf{0.76\pm 0.009}$}
&
\multicolumn{1}{l|}{$\mathbf{ 0.85\pm 0.001}$}            & \multicolumn{1}{l|}{$\mathbf{ 0.82\pm 0.007}$}           & \multicolumn{1}{l}{$\mathbf{ 0.83\pm 0.006}$}
\\ \specialrule{.1em}{.05em}{.05em}
\end{tabular}
}
\vspace{2ex}
\caption{Classification Results: these are the full numeric values (averages and standard deviations computed from three runs) for the graphs shown in \Cref{fig:Classification}.}
\label{tab:full-classification-results}
\end{table*}

\section{Setup/Implementation Details}
\label{app:setup}
The NYT dataset is available through the LDC and a research license, and the Wikipedia dataset is publicly available under a CC BY-NC 4.0 license. %
For both datasets, the token vocabulary size is 150k and the number of unique semantic frames is 600. %
In line with previous research~\citep{rezaee2021event}, for the documents with more than 5 events, we used the first 5 events that had frames. %
Each model was trained using a single GPU (an RTX 2080 TI, TITAN RTX, or a Quadro 8000). %
Each RevUp model took approximately 12 hours to train.

We noticed that the observability of side knowledge can affect the relative importance of the loss terms in \cref{eq:overall-objective}. Given a mini batch of training data, we define $\hat{\epsilon}$ as an approximation of $\epsilon$. To do so, we calculate the number of observed frames over total frames. In all cases, high importance placed on the regularization term $\mathcal{L}_t$ (\cref{eq:t-regularization}) was important to help prevent overfitting. When frames were frequently observed ($\hat{\epsilon} \geq 0.5$), the model needed to rely more heavily on the updating phase ($\mathcal{L}_{\mut}$, \cref{eq:condMutApprox}) and the regularization on $z$ ($\mathcal{L}_z$, \cref{eq:z-regularization}) was less important; specifically, we found $\alpha=0.3$, $\beta=1\mathrm{e}-6$ and $\zeta=0.7$ to be effective. %
Otherwise, we found the model needed to place small but relatively equal weight on the updating phase and $z$ regularization components; specifically, we found $\alpha=0.1$, $\beta=0.2$ and $\zeta=1.0$ to be effective. We obtained these values via experimentation on dev data.

Following SSDVAE and HAQAE models, we use pre-trained Glove 300 embeddings to represent tokens and used gradient clipping at 5.0 to prevent exploding gradients. We use a two-layer of bi-directional GRU for the encoder and a two-layer uni-directional GRU for the decoder
(with a hidden dimension of 512 for both). We used early stopping (lack of validation
performance improvement for 3 iterations).

\end{document}